\documentclass[10pt,twocolumn,letterpaper]{article}

\usepackage{cvpr}      

\usepackage{multirow}
\usepackage{array}
\usepackage{pifont}
\usepackage{subcaption}
\newcommand{\cmark}{\ding{51}}%
\newcommand{\xmark}{\ding{55}}%

\definecolor{cvprblue}{rgb}{0.21,0.49,0.74}
\usepackage[pagebackref,breaklinks,colorlinks,allcolors=cvprblue]{hyperref}

\def\paperID{8908} 
\def\confName{CVPR}
\def\confYear{2025}

\title{Is `Right' Right? Enhancing Object Orientation Understanding in Multimodal Large Language Models through Egocentric Instruction Tuning}
\author{Ji Hyeok Jung$^{1}$\quad Eun Tae Kim$^{1}$\quad Seoyeon Kim$^{1}$\quad Joo Ho Lee$^{1}$\quad Bumsoo Kim$^{2,}$\thanks{Corresponding authors.}\quad Buru Chang$^{3,}$\footnotemark[1]\\
$^1$Sogang University\quad $^2$Chungang University\quad $^3$Korea University\\
{\tt\small \{ji9759,untae0122,ksy02031,jhleecs\}@sogang.ac.kr}\quad
{\tt\small bumsoo@cau.ac.kr}\quad{\tt\small buru\_chang@korea.ac.kr}
} 

\begin{document}
\maketitle
\begin{abstract}
Multimodal large language models (MLLMs) act as essential interfaces, connecting humans with AI technologies in multimodal applications.
However, current MLLMs face challenges in accurately interpreting object orientation in images due to inconsistent orientation annotations in training data, hindering the development of a coherent orientation understanding.
To overcome this, we propose egocentric instruction tuning, which aligns MLLMs' orientation understanding with the user’s perspective, based on a consistent annotation standard derived from the user’s egocentric viewpoint. 
We first generate egocentric instruction data that leverages MLLMs' ability to recognize object details and applies prior knowledge for orientation understanding. 
Using this data, we perform instruction tuning to enhance the model’s capability for accurate orientation interpretation.
In addition, we introduce EgoOrientBench, a benchmark that evaluates MLLMs' orientation understanding across three tasks using images collected from diverse domains.
Experimental results on this benchmark show that egocentric instruction tuning significantly improves orientation understanding without compromising overall MLLM performance. 
The instruction data and benchmark dataset are available on our project page at \url{https://github.com/jhCOR/EgoOrientBench}.
\end{abstract}
\section{Introduction}\label{sec:1_introduction}
Multimodal large language models (MLLMs)~\cite{dai2023instructblipgeneralpurposevisionlanguagemodels,zhu2024minigpt,liu2024visual} can serve as interfaces connecting humans with AI technologies designed to perform multimodal tasks that require both image and text understanding.
For instance, MLLMs are useful for operating autonomous systems~\cite{shao2024lmdrive,wei2024editable}, manipulating robots~\cite{gao2023physically,guan2024loc}, and facilitating communication with AI assistants on AR devices~\cite{wang2023holoassist,pei2024autonomous,mu2024embodiedgpt}.
To ensure consistent interaction, it is essential for MLLMs to exhibit behaviors that align accurately with human intentions.

\begin{figure}[t]
  \centering
  \includegraphics[width=\columnwidth]{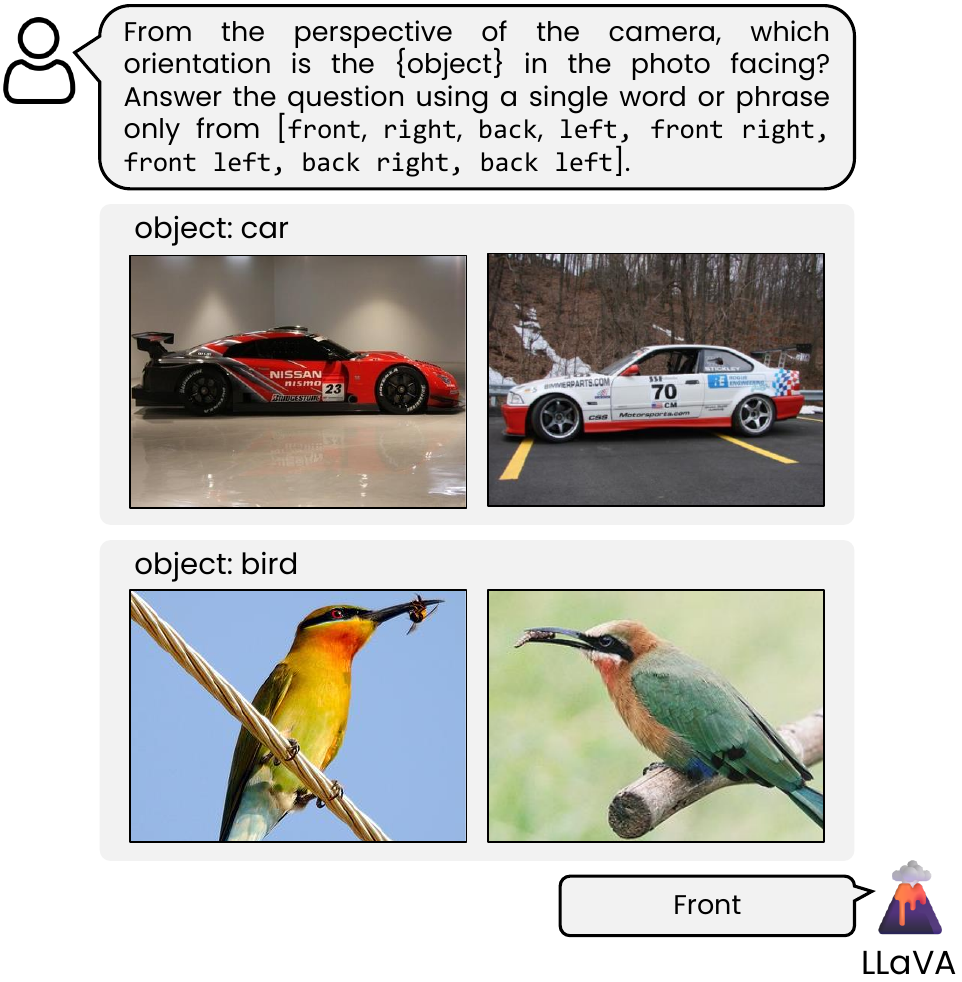}
  \caption{
Examples of LLaVA's~\cite{liu2024visual} responses to prompts asking about the orientation of objects in an image. Current MLLMs show a significant lack of understanding when interpreting the orientation of given objects.}
  \label{fig:1_motivation}
  \vspace{-1em}
\end{figure}
\begin{figure*}[t]
  \centering
  \includegraphics[width=\textwidth]{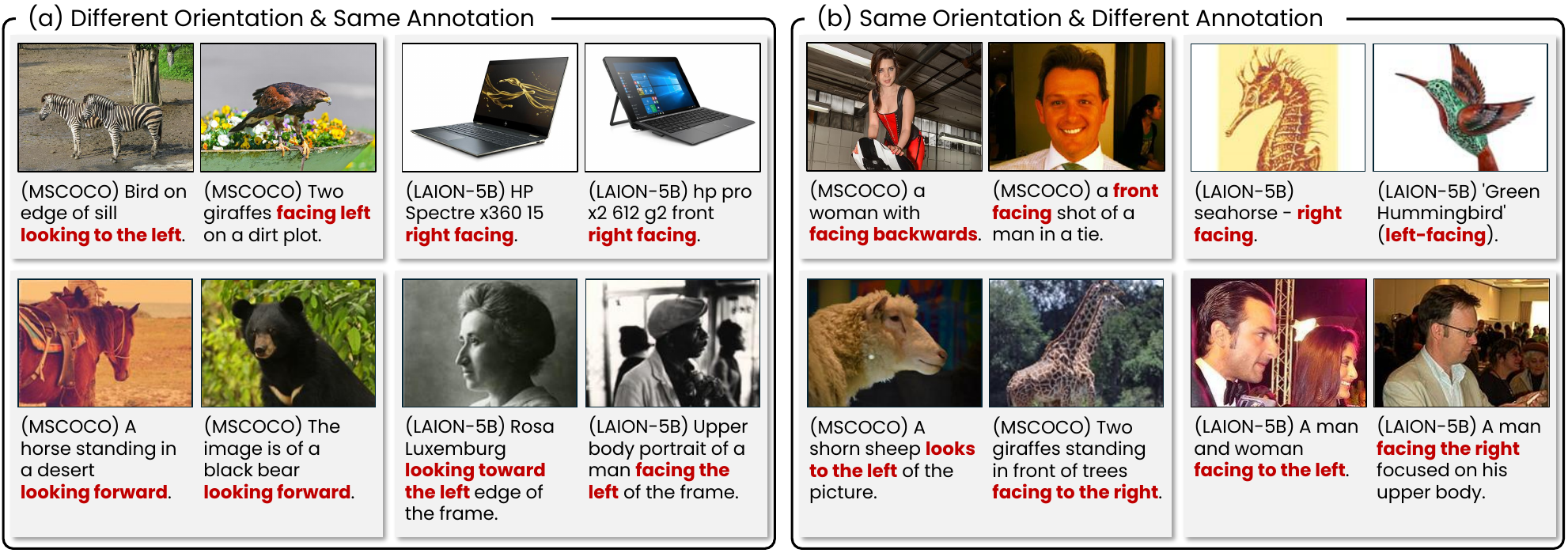}
  \caption{
Examples of inconsistent annotations for object orientation. In image-text pairs used for training MLLMs, such as those in MSCOCO~\cite{lin2014microsoft} and LAION-5B~\cite{schuhmann2022laion}, annotations for object orientation lack consistency. 
For instance, (a) objects facing different orientations may be annotated as facing the same orientation, or conversely, (b) objects facing the same orientation may be annotated differently. This variation arises because, without a standardized guideline for object orientation, annotations can vary depending on individual perspectives or cultural backgrounds~\cite{levinson1996frames}. Our study aims to improve MLLM’s understanding of object orientation by aligning it with the user's egocentric perspective through instruction tuning.}
  \label{fig:2_dataset_example}
\end{figure*}

However, MLLMs’ limited understanding of object orientation complicates effective communication. 
As shown in Figure~\ref{fig:1_motivation}, current MLLMs struggle to accurately interpret the orientation an object is facing. 
This limitation can lead to errors in autonomous vehicle navigation, incorrect robot operations, and even severe accidents, creating a significant barrier to deploying MLLMs in real-world applications.


In this study, we identify inconsistent annotations in training data—without standardized guidelines—as a major factor hindering accurate orientation understanding.
Unlike humans, who understand object orientation through an egocentric reference frame based on their physical bodies~\cite{gramann2010human,zaehle2007neural,vallar1999fronto}, MLLMs, lacking a physical body, rely solely on training data to develop orientation understanding.
However, as shown in Figure~\ref{fig:2_dataset_example}, datasets used to train MLLMs like MS-COCO~\cite{lin2014microsoft} and LAION-5B~\cite{schuhmann2022laion} contain inconsistent annotations for object orientation, sometimes based on the observer’s perspective and other times on the object’s perspective.
Such inconsistent annotations for objects make it difficult for MLLMs to understand object orientation, leading to behaviors misaligned with user intentions.
Since annotations can vary depending on individual or cultural backgrounds~\cite{levinson1996frames}, annotation inconsistencies are inevitable without a consistent annotation standard that considers real-world use cases for MLLMs.
%

To address this issue, we propose \textit{Egocentric Instruction Tuning}, which aligns MLLMs' understanding of object direction with the user’s perspective. 
Considering practical application scenarios where users and MLLMs interact from a unified viewpoint, this tuning method adjusts the MLLM’s orientation understanding to align with the user’s egocentric perspective.
To implement this, we manually annotate object orientation based on an egocentric perspective using ImageNet data~\cite{deng2009imagenet}, then generate egocentric instruction data from these annotations. 
This process leverages MLLM’s ability to recognize image details and the LLM’s prior knowledge to associate these details with object orientation. 
By applying instruction tuning with this data, we improve alignment with human perception of object orientation without compromising general performance.

Furthermore, we introduce \textit{EgoOrientBench} (\underline{Ego}centric Object \underline{Orient}ation Understanding \underline{Bench}mark), a benchmark to evaluate MLLM object orientation comprehension.
Existing datasets~\cite{tong2024eyes} for assessing this capability are limited to small collections with only a few dozen samples, restricting comprehensive evaluation.
Our benchmark includes three tasks of varying complexity to enable a thorough assessment of object orientation comprehension. 
Evaluation data is collected from diverse domains—including photographs, painting, drawing, and 3D rendering—to reflect a wide range of application use cases.
Experimental results on this benchmark demonstrate that aligning object orientation understanding in MLLMs with the user perspective through egocentric instruction tuning enhances applicability in real-world scenarios.

\noindent
\textbf{Contributions. } Our contributions can be summarized as follows:
(1) We introduce a new task to enhance and evaluate MLLMs' understanding of object orientation from the user’s egocentric perspective.
(2) We propose \textit{egocentric instruction tuning}, a method that aligns MLLM’s understanding of object orientation with the user’s egocentric perspective, and provide the necessary data for this approach. This method enhances orientation understanding without compromising general performance.
(3) We introduce \textit{EgoOrientBench}, a benchmark designed for comprehensive evaluation of object orientation understanding in MLLMs.

\section{Related Work}\label{sec:2_related_work}
\subsection{Multimodal Large Language Model}\label{subsec:2_1_multimodal_large_language_model}
Recently, numerous studies have concentrated on developing MLLMs that can effectively interpret both visual and textual data. 
For example, Flamingo~\cite{alayrac2022flamingo} leverages visual and textual inputs as prompts, achieving notable few-shot performance in visual question answering. 
Models like LLaVA~\cite{liu2024visual} and MiniGPT4-v2~\cite{zhu2024minigpt} apply visual instruction tuning to align MLLM behavior more closely with user intentions.
Moreover, methods such as VisionLLM~\cite{wang2024visionllm}, KOSMOS-2~\cite{peng2023kosmos}, and Qwen-VL~\cite{bai2023qwen} have demonstrated superior performance in visual grounding tasks.
InternVL~\cite{chen2024internvl} underscores the significance of scaling the visual encoder in proportion to the LLM’s size, while PaLM-E~\cite{driess2023palm} and EmbodiedGPT~\cite{mu2024embodiedgpt} reveal the promising potential of MLLMs for embodied applications.
Additionally, commercial APIs, including GPT-4~\cite{achiam2023gpt}, Claude 3.5~\cite{anthropic2024claude}, and Gemini-1.5~\cite{team5gemini}, are actively utilizing MLLMs across a range of applications.

Nevertheless, previous research~\cite{tong2024eyes,gaur2024detect} has identified that current MLLMs struggle with accurately understanding object orientation.
In this study, we highlight that this limitation arises from the inconsistent annotations regarding object orientation in the image-text pair data used for MLLM training. 
To address this issue, we focus on developing an egocentric instruction tuning method that aligns MLLMs’ understanding of object orientation with human perception, thereby enhancing orientation understanding in user-centered applications.

\subsection{Object Orientation Understanding}\label{subsec:2_2_object_orientaton_understanding}
Research on object orientation understanding is actively pursued across various fields. Pose estimation~\cite{ozuysal2009pose,di2022gpv} is a fundamental task in interpreting object orientation in images. 
In autonomous driving, for instance, understanding pedestrian body orientation aids in predicting movement paths and reducing accident risk~\cite{enzweiler2010integrated,dollar2011pedestrian,gandhi2008image}. Recent studies have explored enhancing object orientation comprehension by incorporating viewpoint information.
For example, DREAMFUSION~\cite{pooledreamfusion} uses orientation data from multiple viewpoints to create enriched object descriptions, while EgoExoLearn~\cite{huang2024egoexolearn} focuses on harmonizing orientation interpretation across varying perspectives.

More recently, MMVP~\cite{tong2024eyes} has underscored the need for object orientation understanding within MLLMs. 
To evaluate encoder representations, MMVP introduces a simple evaluation method and dataset that measure MLLMs’ ability to interpret object orientation. 
However, this approach is limited by a small sample size of real-image data, which restricts comprehensive assessment across diverse image domains and various orientation complexities.
To address this limitation, we propose a large-scale benchmark specifically designed to evaluate MLLMs' object orientation understanding across a broad range of domains and conditions, facilitating a more robust analysis of MLLM capabilities.
\section{Egocentric Instruction Tuning}\label{sec:3_egocentric_instruction_tuning}
\subsection{Egocentric Annotation}\label{subsec:3_1_egocentric_annotation}
\textbf{Motivation.}
Inconsistent annotations hinder MLLMs' ability to develop an understanding of object orientation. 
Therefore, consistent annotations should be used in MLLM training, which requires a standardized annotation guideline.
Meanwhile, with the rise of embodied AI, numerous studies~\cite{chen2024egocentric,grauman2022ego4d,xu2024retrieval} have explored using AI technologies from a user's egocentric perspective. 
For instance, over ten egocentric video datasets, such as EgoExoLearn~\cite{huang2024egoexolearn}, EgoCoT~\cite{mu2024embodiedgpt}, and EgoSchema~\cite{mangalam2023egoschema}, have been introduced in the past year. 
Motivated by this trend, we standardize annotations based on the user’s egocentric viewpoint to address the issue of inconsistent object orientation annotations. 
This standardization enhances the applicability of MLLMs in real-world, user-centered applications.

\noindent
\textbf{Orientation Classes.}
We construct the egocentric instruction data using a consistent annotation rule. 
This rule categorizes object orientation into eight distinct egocentric classes (see Figure~\ref{fig:4_benchmark_data}):
\begin{itemize}
    \item \textit{\textbf{Front}}: the object faces the user (or camera).
    \item \textit{\textbf{Back}}: the object is turned away, facing the opposite direction from the user (or camera).
    \item \textit{\textbf{Left}}: the object is oriented to the left of the user.
    \item \textit{\textbf{Right}}: the object is oriented to the right of the user.
    \item \textit{\textbf{Front-Left}}, \textit{\textbf{Front-Right}}: the object faces the user but is angled toward the left (or right).
    \item \textit{\textbf{Back-Left}}, \textit{\textbf{Back-Right}}: the object faces away from the user but is angled toward the left (or right).
\end{itemize}
This classification scheme provides a structured framework to interpret object orientation consistently from a user-centered perspective.

\subsection{Egocentric Instruction Data}\label{subsec:3_2_egocentric_instruction_data_generation}
To create the egocentric instruction dataset, we begin by manually annotating object orientations using the ImageNet data~\cite{deng2009imagenet}.
We then generate LLaVA-style instruction data~\cite{liu2024visual} leveraging MLLM’s ability to recognize image details and the LLM's prior knowledge to associate these details with object orientation. 
Using this dataset, we perform instruction tuning to enhance the MLLMs' ability to interpret object orientation while maintaining their general response generation capabilities.

\begin{figure*}[t]
  \centering
  \includegraphics[width=\textwidth]{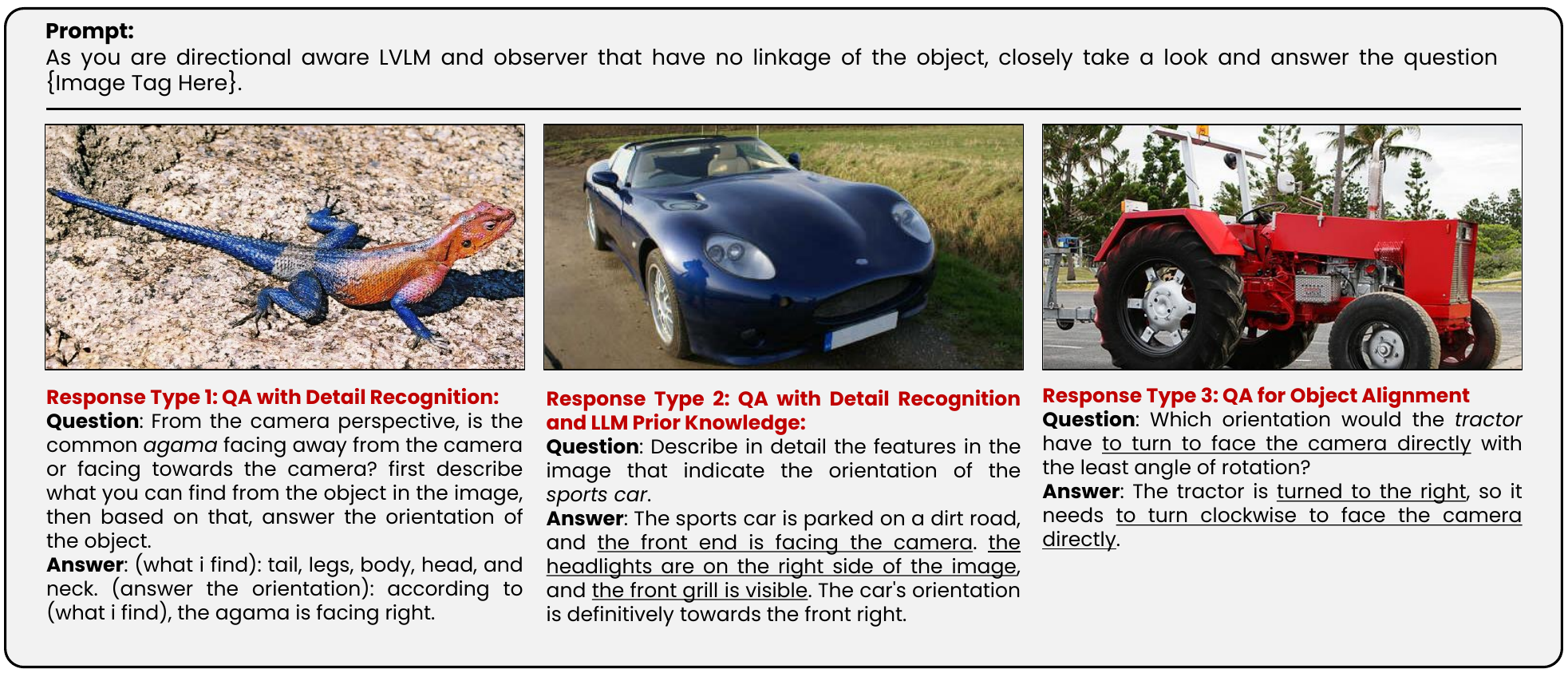}
  \caption{
An example of egocentric instruction data designed to enhance MLLMs' understanding of object orientation. This data leverages the model's intrinsic ability to recognize object details (Response Type 1) and the LLM's prior knowledge to link these details to specific orientations (Response Type 2). Furthermore, by engaging in object alignment tasks that require understanding the relationships between different orientations (Response Type 3), MLLMs’ comprehension of object orientation is further improved.}
  \label{fig:3_egocentric_instruction_data}
\end{figure*}

\noindent
\textbf{Manual Data Collection.
}
Among the datasets used for training MLLMs, none provide egocentric annotations for object orientation.
To address this gap, we initiate the first effort to manually annotate object orientation in an egocentric context, creating a unique dataset specifically designed to enhance MLLMs’ understanding of object orientation. 
We focus initially on single-object scenarios to establish a foundation for further exploration and development.
Using ImageNet data~\cite{deng2009imagenet} and the annotation scheme detailed in Section~\ref{subsec:3_1_egocentric_annotation}, we classify objects into eight distinct orientations, ensuring clarity and minimizing ambiguity in the dataset.
Dataset statistics are summarized in Appendix~\ref{subsec:a_2_data_statistics}.

\noindent
\textbf{Instruction Data Generation.}
Using manually annotated orientation information, we create egocentric instruction tuning data inspired by LLaVA data~\cite{liu2024visual} to align object orientation understanding with the user's egocentric perspective while preserving MLLMs' general response generation capabilities.
Instead of directly answering an object’s orientation, we design three response types to enhance orientation understanding by leveraging the model’s ability to recognize image details and the LLM’s prior knowledge: 
\begin{itemize} 
    \item \textbf{Response Type 1}: Utilizes MLLM's detail recognition abilities. 
    The model identifies specific image details, such as a car’s taillights or an elephant’s tail, that provide clues about orientation, guiding MLLM to interpret the object’s orientation more accurately. 
    \item \textbf{Response Type 2}: Draws on the LLM's prior knowledge essential for understanding orientation, instructing the model to describe orientation based on key details. 
    For example, the LLM knows that a human face's orientation depends on the nose position, while a lizard's orientation relies on the head-tail alignment. 
    This type links prior knowledge with image details, enabling MLLM to describe the object and conclude its orientation. 
    \item \textbf{Response Type 3}: Adapts from robot manipulation tasks to improve MLLM's understanding of orientation relationships. 
    Aligning an object to a target orientation requires grasping the relation between its current and target orientations. 
    For instance, rotating an object facing front-left counterclockwise aligns it to front. 
    This type enhances MLLM's understanding of orientation relationships from an egocentric perspective. 
\end{itemize} 
In Section~\ref{subsec:5_3_analysis}, we will conduct ablation tests to validate the contribution of each response type to improved object orientation understanding.

\subsection{Instruction Tuning}\label{subsec:3_3_instruction_tuning}
We conduct instruction tuning on the MLLM using the previously generated egocentric instruction data. 
During this process, we keep the visual encoder frozen and apply supervised fine-tuning to both the LLM and the bridge layer. 
To increase training efficiency, we employ LoRA~\cite{hulora} for parameter-efficient fine-tuning.

\section{\textit{EgoOrientBench}: \underline{Ego}centric \underline{Orient}ation Understanding \underline{Bench}mark}\label{sec:4_egocentric_directional_understanding_benchmark}
\subsection{Data Collection.}\label{subsec:4_1_data_collection}
\begin{figure}[t]
  \centering
  \includegraphics[width=\columnwidth]{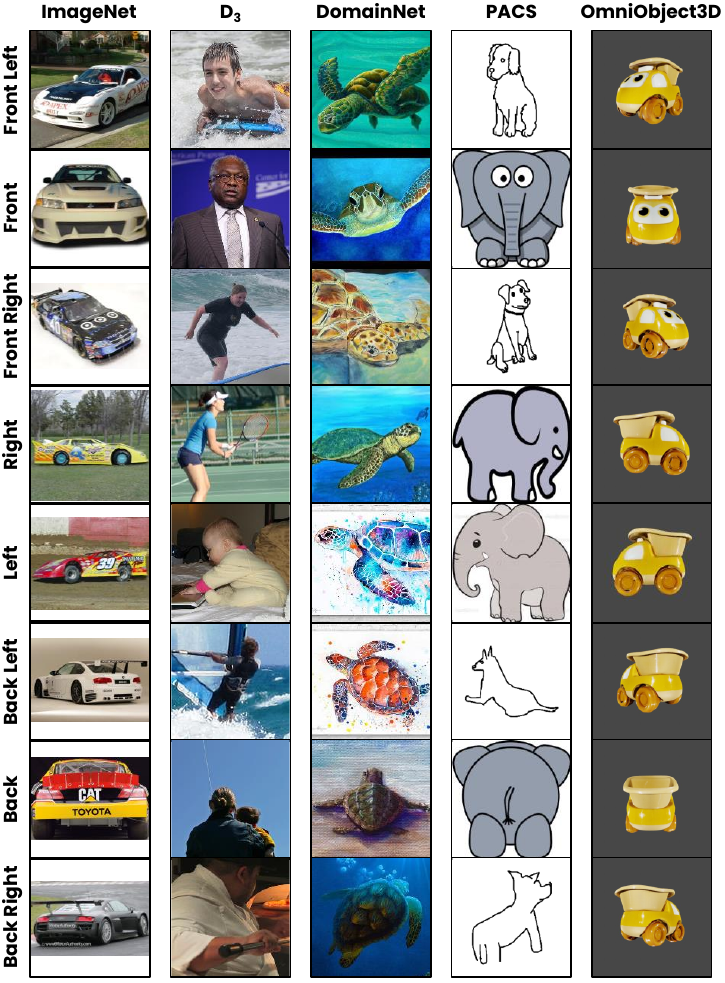}
  \caption{
Our benchmark data examples. Our benchmark consists of data collected from various image domains to assess the applicability of MLLMs in terms of orientation understanding. The collected data is annotated across eight orientation classes.}
\label{fig:4_benchmark_data}
\vspace{-1em}
\end{figure}

MLLMs, known for their versatile capabilities, can be applied across various domains. 
To assess MLLMs' ability to recognize object orientation in diverse domains, we introduce \textit{EgoOrientBench}, a large-scale benchmark dataset designed for orientation understanding from a user-centered, egocentric perspective.
The dataset is constructed from five distinct image datasets, spanning various domains such as drawings, paintings (from PACS and DomainNet), and 3D-rendered images (from OmniObject3D).
\begin{itemize}
    \item \textbf{ImageNet}~\cite{deng2009imagenet}: Provides real images of single objects across 1,000 object classes.
    \item \textbf{D$_3$}~\cite{gaur2024detect}: Offers a small-scale data designed for a comprehensive assessment of MLLM's object understanding capabilities, including several dozen examples that evaluate object orientation in real images.
    \item \textbf{DomainNet}~\cite{peng2019moment}: Includes cross-domain evaluation images, featuring single-object representations in sketches, paintings, and clip art.
    \item \textbf{PACS}~\cite{li2017deeper}: Contains images from multiple domains, including real photos, art paintings, cartoons, and sketches. 
    \item \textbf{OmniObject3D}~\cite{wu2023omniobject3d}: Supplies mesh data for 3D rendering, enabling the capture of object orientations by adjusting camera angles.
\end{itemize}
We annotate the data across these sources according to the eight orientation classes outlined in Section~\ref{subsec:3_1_egocentric_annotation}. 
For the ImageNet dataset, we collected 50 images per class across the eight orientations. 
In the D$_3$, DomainNet, and PACS datasets, images for the \textit{Back}, \textit{Back-Left}, and \textit{Back-Right} orientations are sparse, resulting in imbalanced annotations. 
OmniObject3D, which allows rendering of an object in all eight orientations by adjusting the camera angle, enable balanced annotation across orientations. 
Figure~\ref{fig:4_benchmark_data} provides examples of this benchmark evaluation data, and detailed statistics are presented in Appendix~\ref{subsec:a_2_data_statistics}.

\subsection{Task Description.}\label{subsec:4_2_task_description}
We design three tasks to assess MLLMs' object orientation understanding at various levels 
\begin{itemize}
    \item \textit{\textbf{Choose}}: A simple classification task where the model selects an object’s orientation from eight multiple-choice options to assess basic orientation understanding.
    \item \textit{\textbf{Verify}}: A binary classification task where the model determines whether an object’s orientation matches a given query, instead of choosing from set options.
    \item \textit{\textbf{Freeform}}: Unlike the previous tasks, this task involves generating a descriptive response for an object’s orientation, rather than selecting from options. 
    GPT-4 then assesses the alignment between the generated description and the true orientation.
\end{itemize}
Each task targets specific aspects of orientation understanding, while offsetting each other’s limitations. 
For example, \textit{Choose} is effective for basic orientation comprehension but limited by the division of continuous orientation into discrete classes. 
To overcome this, \textit{Verify} tests the model's ability to match the given orientation in the query to an object in an image. 
Finally, since descriptive vocabulary can vary, \textit{Freeform} provides flexible evaluation by generating descriptive orientation responses, matched to the true answer using GPT-4. 
Task prompts of EgoOrientBench are detailed in Appendix~\ref{subsec:a_3_task_data_details}.

\section{Experiments}\label{sec:5_experiments}
\subsection{Experimental Setup}\label{subsec:5_1_experimental_setup}
\textbf{Backbone Models.}
To measure the current object orientation understanding of MLLMs and verify the effectiveness of our egocentric instruction tuning, we evaluate the following three MLLMs in EgoOrientBench: LLaVA 1.5 (w/ Vicuna-7B)~\cite{liu2024visual}, mPLUG-Owl2 (w/ LLaMA2-7B)~\cite{ye2024mplug}, and InternVL2 (4B)~\cite{chen2024internvl}. 
Additionally, we also conduct evaluations for commercial MLLMs, GPT-4o (\texttt{gpt-4o-2024-08-06})~\cite{achiam2023gpt} and Gemini-1.5 (\texttt{Flash-8B})~\cite{team5gemini} using their API.

\noindent
\textbf{Benchmarks.}
To verify the effectiveness of egocentric instruction tuning, we conduct experiments in \textbf{\textit{EgoOrientBench}} to assess MLLMs' understanding of object orientation. 
Additionally, to determine whether egocentric instruction tuning enhances object orientation understanding without compromising general response generation, we evaluate the models' generalization capabilities using four benchmarks: \textbf{\textit{MME}}~\cite{fu2023mme}, \textbf{\textit{MMStar}}~\cite{chen2024we}, \textbf{\textit{MMMU}}~\cite{yue2024mmmu}, and \textbf{\textit{POPE}}~\cite{li2023evaluating}.

\begin{table*}[t]
\centering
\small
\resizebox{\textwidth}{!}{%
\begin{tabular}{ll|ccccc|c|ccccc|c|ccccc|c} 
 \toprule
 \multicolumn{2}{c|}{Method} & \multicolumn{6}{c|}{Choose} & \multicolumn{6}{c|}{Verify} & \multicolumn{6}{c}{Freeform} \\ 
 \midrule
\multicolumn{1}{l}{Backbone}&\multicolumn{1}{l|}{Method}&IN&D$_{3}$&DN&PA&3D&Avg.&IN&D$_{3}$&DN&PA&3D&Avg.&IN&D$_{3}$&DN&PA&3D&Avg.\\
\midrule
- & \textit{Popular} & 12.5 & 25.0 & 27.7 & 30.7 & 12.5 & 20.3 & 50.0 & 50.0 & 50.0 & 50.0 & 50.0 & 50.0 & \multicolumn{6}{c}{-}\\
\midrule

\multirow{2}{*}{LLaVA-1.5 (7B)} &Zero-shot & 12.2 & 22.9 & 24.0 & 23.9&12.5&17.9&52.2&49.3&51.3&51.2&52.3&51.8&14.7&15.1&21.3&20.5&15.5&17.9\\ 
 &Ours & \textbf{42.4} & \textbf{37.2} & \textbf{42.4} & \textbf{40.8} & \textbf{25.2} & \textbf{33.7} & \textbf{62.2} & \textbf{56.0} & \textbf{61.6} & \textbf{58.3} & \textbf{56.7} & \textbf{58.3} &\textbf{65.5}&\textbf{50.0}&\textbf{72.3}&\textbf{67.2}&\textbf{43.3}&\textbf{56.5}\\ 
\midrule

\multirow{2}{*}{mPLUG-Owl2 (7B)} & Zero-shot & 14.2  & 23.4 & 18.4 & 15.9 & 12.8 & 15.0  & 54.4 & 49.5 & 50.2 & 46.4 &50.9&49.8&19.3&28.3&26.6 &26.6 & 28.3&20.7\\ 
 & Ours & \textbf{34.5} & \textbf{34.0} & \textbf{33.0} & \textbf{37.1} & \textbf{21.2} & \textbf{28.5} & \textbf{67.1} & \textbf{62.8} & \textbf{64.2} & \textbf{65.4} & \textbf{57.6} & \textbf{61.5} &\textbf{41.5} &\textbf{40.1}&\textbf{42.5}&\textbf{48.1}& \textbf{28.4}&\textbf{37.1}\\ 
\midrule

\multirow{2}{*}{InternVL2 (4B)} & Zero-shot&15.4&15.3&21.1&15.7&15.2&16.5&54.1&56.6&63.0&59.5&54.9&57.7&22.0&39.5&53.1&52.3&28.9&39.6 \\ 
 & Ours &\textbf{18.7}&\textbf{30.9}&\textbf{42.0}&\textbf{50.3}&\textbf{18.3}&\textbf{31.4}&\textbf{58.8}&\textbf{58.5}&\textbf{68.2}&\textbf{67.0}&\textbf{56.0}&\textbf{61.4}&\textbf{41.2}&\textbf{43.4}&\textbf{66.8}&\textbf{66.1}&\textbf{31.6}&\textbf{48.2}\\  
\midrule

\multirow{2}{*}{API}
& Gemini-1.5&24.5&25.7&29.3&34.8&30.0&30.7&55.0&53.0&56.1&61.5&56.5&57.5&36.2 &51.3&55.0&61.3&49.3&52.9\\
& GPT-4o &36.0&44.7&47.5& 55.8 & 31.1 &41.1 &75.7& 77.0&79.8&77.8&73.8&76.2&58.3&63.2&86.8&87.6&59.8&72.3\\

\bottomrule
\end{tabular}%
}

\caption{EgoOrientBench results (\%). We report the results for three tasks (Choose, Verify, Freeform) on five datasets, ImageNet (IN)~\cite{deng2009imagenet}, D$_3$~\cite{gaur2024detect}, DomainNet (DN)~\cite{peng2019moment}, PACS (PA)~\cite{li2017deeper}, and OmniObject3D (3D)~\cite{wu2023omniobject3d}. \textit{Popular} indicates the performance achieved by uniformly responding with the most major class in each task, serving as the most basic baseline performance on imbalanced data.}

\label{tab:1_benchmark_results}
\vspace{-0.5em}
\end{table*}

\noindent
\textbf{Implementation Details.}All models are trained on a single A6000 48GB GPU for 3 epochs. 
To maintain general response generation capabilities, the visual encoder remains frozen across all models. 
In our implementation, we train the LLaVA-1.5, mPLUG-Owl2, and InternVL2 models following the hyperparameters specified in the original papers. 
All models are trained using the AdamW optimizer~\cite{loshchilov2018decoupled}.
The learning rates are set as follows: $2e$-$4$ for LLaVA-1.5, $1e$-$4$ for mPLUG-Owl2, and $4e$-$5$ for InternVL2. 
A warm-up ratio of $0.3$ and a cosine learning rate scheduler are applied to all models. 
Additionally, weight decay of $0.05$ is applied exclusively to InternVL2.
We use LoRA~\cite{hulora} for parameter-efficient fine-tuning, with settings of $r$=$128$ and $\alpha$=$256$ for LLaVA-1.5 and mPLUG-Owl2, and $r$=$16$ and $\alpha$=$32$ for InternVL2.
Batch sizes are set according to each model's memory needs: LLaVA-1.5 uses a batch size of 4 with gradient accumulation of 4, mPLUG-Owl2 uses a batch size of 2 with gradient accumulation of 16, and MiniGPT-V2 uses a batch size of 1. 
During inference, tasks are validated one at a time without batching.

\subsection{Experimental Result}\label{subsec:5_2_experimental_result}
\textbf{Results on EgoOrientBench.}
Table~\ref{tab:1_benchmark_results} presents the evaluation results on EgoOrientBench, showing that our egocentric instruction tuning significantly enhances MLLMs' understanding of object orientation across all tasks. 
Specifically, all zero-shot MLLMs perform worse on the \textit{Choose} task compared to the \textit{Popular} method, which uses the most common orientation in each dataset as the default response. 
This performance gap stems from a strong orientation bias in current MLLMs on the \textit{Choose} task, as analyzed in Section~\ref{subsec:5_3_analysis}. 
The improvements achieved with our method suggest that it effectively mitigates this bias, resulting in better orientation comprehension.
Although the commercial MLLM GPT-4o achieves the highest overall performance, our improved versions of LLaVA-1.5 and InternVL2 outperform another commercial model, Gemini-1.5, in all tasks.

Note that this enhancement occurs even with the visual encoder frozen, without additional training.
While previous studies have attributed MLLMs' limitations in object orientation understanding to the visual encoder's inability to represent orientation information~\cite{tong2024eyes,gaur2024detect}, our benchmark experiments demonstrate that substantial improvements can be achieved by aligning MLLM orientation interpretation with the user’s egocentric perspective. 

\begin{table}[t]
\centering
\small
\resizebox{\columnwidth}{!}{%
\begin{tabular}{ll|cccc}
\toprule
\multicolumn{2}{c|}{Method} & \multirow{2}{*}{MME} & \multirow{2}{*}{MMStar} & \multirow{2}{*}{MMMU} & \multirow{2}{*}{POPE} \\ 
\multicolumn{1}{l}{Backbone}&\multicolumn{1}{l|}{Method}& & &&\\ \midrule
\multirow{2}{*}{LLaVA 1.5} & Zero-shot & 1792.8 & 34.67 & 35.11 & 82.03 \\
                            & Ours      & 1752.8 & 35.87 {\scriptsize } & 34.44 {\scriptsize } & 88.36 {\scriptsize } \\
\midrule
\multirow{2}{*}{mPLUG-Owl2} & Zero-shot & 1706.3 & 34.33 & 37.55 & 86.16 \\
                            & Ours      & 1727.3 & 35.27 {\scriptsize } & 38.55 {\scriptsize } & 85.60 {\scriptsize } \\
\midrule
\multirow{2}{*}{InternVL2-4B} & Zero-shot & 2088.7 & 54.26 & 47.22 & 85.91 \\
                            & Ours      & 2045.9 & 53.13 {\scriptsize } & 48.00 {\scriptsize } & 85.56 {\scriptsize } \\
\bottomrule
\end{tabular}%
}
\caption{Experimental results evaluating the general performance of MLLMs on the MME~\cite{fu2023mme}, MMStar~\cite{chen2024we}, MMMU~\cite{yue2024mmmu}, and POPE~\cite{li2023evaluating} before and after egocentric instruction tuning.}
\label{tab:2_combined_results}
\end{table}

\noindent
\textbf{Results on general benchmarks.}
Table~\ref{tab:2_combined_results} presents the experimental results on the general performance benchmarks. 
The results indicate that performance remains stable after applying our egocentric instruction tuning, with no significant degradation compared to pre-tuning levels. 
Overall, the three models exhibit largely comparable performance across the four benchmarks. 
In particular, LLaVA 1.5 shows a significant improvement in \textit{POPE}, 
while mPLUG-Owl2 demonstrates considerable gains in \textit{MME}. 
Similarly, InternVL2 maintains stable or slightly improved performance without noticeable degradation across most benchmarks.
 
These findings demonstrate that our instruction tuning effectively aligns MLLM’s object orientation understanding with the user’s perspective while maintaining the model’s overall response generation capabilities.

\subsection{Analysis}\label{subsec:5_3_analysis}
\begin{figure}[t]
\centering
\begin{subfigure}{\columnwidth}
  \centering
\includegraphics[width=\columnwidth]{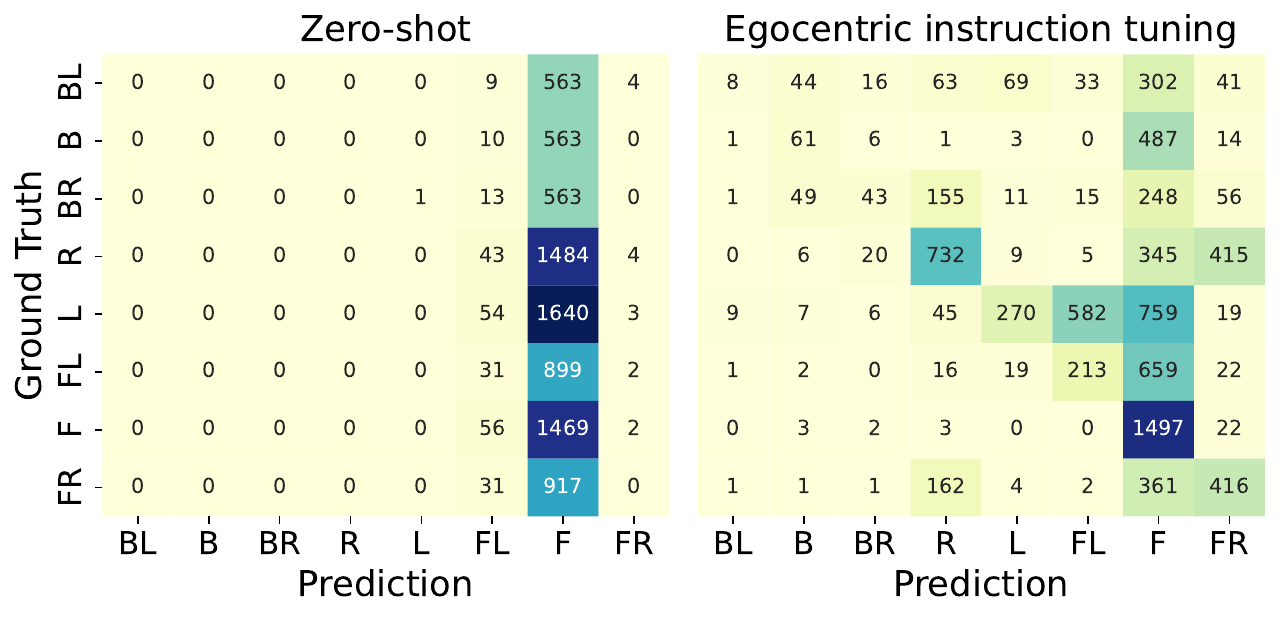}
  \caption{LLaVA}
\label{fig:5_1_llava_heatmap}
\end{subfigure}%
\vspace{0.0em}
\begin{subfigure}{\columnwidth}
  \centering
\includegraphics[width=\columnwidth]{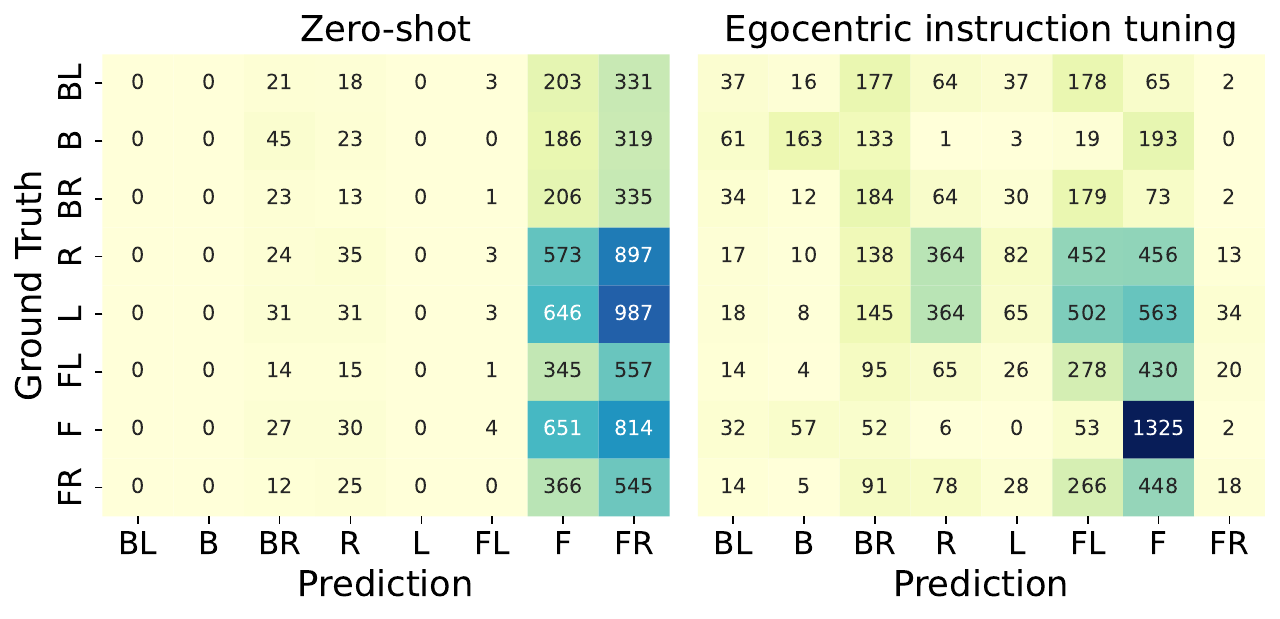}
  \caption{mPLUG-Owl2}
  \label{fig:5_2_owl_heatmap}
\end{subfigure}%
\caption{Confusion matrix for the \textit{Choose} task with LLaVA and mPLUG-Owl2. Zero-shot MLLMs show extreme bias toward the \textit{Front} class (F) (or the \textit{Front Right} (FR) class). Our proposed egocentric instruction tuning relieves this bias and enhances the understanding of object orientation.}
\label{fig:5_confusion_matrix}
  \vspace{-0.5em}
\end{figure}

\textbf{Confusion Matrix.}
We analyze error cases before and after egocentric instruction tuning by examining the confusion matrix for the \textit{Choose} task. 
As shown in Figure~\ref{fig:5_confusion_matrix}, LLaVA and mPLUG-Owl2 produce zero-shot predictions that are heavily biased toward the specific orientations, particularly the \textit{Front} (F) and \textit{Front Right} (FR) classes, indicating a significant deficiency in object orientation understanding in publicly available MLLMs. 
We also observe that this bias is mitigated through our proposed egocentric instruction tuning.
Notably, there are several acceptable error cases, such as predicting objects in the \textit{Left} (L) class as \textit{Front Left} (FL) or interpreting \textit{Front Right} (FR) objects as \textit{Front} (F) or \textit{Right} (R) classes. 
This confusion matrix analysis is consistent with the benchmark results in Table~\ref{tab:1_benchmark_results}, further confirming that our tuning method effectively enhances object orientation understanding.
We want to emphasize that our EgoOrientBench provides a foundation for analyzing MLLM behaviors related to object orientation, supporting future research efforts to enhance MLLM orientation comprehension.
The confusion matrix analysis for InternVL2 is provided in Appendix~\ref{subsec:c_1_confusion_matrix}.

\noindent
\textbf{Ablation Test.}
Table~\ref{tab:3_ablation_test} shows the ablation test results for the three response types in our egocentric instruction data. 
Using all three response types yields the highest performance, while removing any type results in a decrease in task accuracy, indicating that each type plays a role in enhancing MLLM’s object orientation understanding. 
Notably, combining Type 1 and Type 2 leads to significant improvements across all tasks, suggesting a strong synergy between them. 
Adding Type 3 further boosts performance, showing that it contributes to orientation understanding in a way that complements Types 1 and 2.
The ablation results for mPLUG-Owl2 and InternVL2 are presented in Appendix~\ref{subsec:c_2_ablation_test}.
\begin{table}[!t]
\centering
\scriptsize
\resizebox{\columnwidth}{!}{%
\begin{tabular}{cccccc}
\toprule
\multicolumn{3}{c}{Response Types} & \multirow{2}{*}{Choose} & \multirow{2}{*}{Verify} & \multirow{2}{*}{Freeform} \\
Type 1& Type 2 & Type 3 & & &  \\\midrule
\cmark & \xmark & \xmark& 18.3&54.7&32.4\\
\xmark & \cmark & \xmark& 18.3&54.5&38.6\\
\xmark & \xmark & \cmark& 30.0&50.3&46.1\\
\midrule

\cmark & \cmark & \xmark& 23.3&59.3&43.4\\
\cmark & \xmark & \cmark& 27.7&56.4&46.6\\
\xmark & \cmark & \cmark& 29.7&54.9&54.6\\
\midrule
\cmark & \cmark & \cmark& 33.7&58.3&56.5\\
\bottomrule
\end{tabular}%
}
\caption{Ablation test with LLaVA-1.5. Each response type contributes to performance improvements across all three tasks.}
\label{tab:3_ablation_test}
\end{table}



\section{Discussion}\label{sec:6_discussion}
\subsection{Applicability in Real-world Scenarios}\label{subsec:6_1_applicability}
\begin{figure}[t]
  \centering
  \includegraphics[width=\columnwidth]{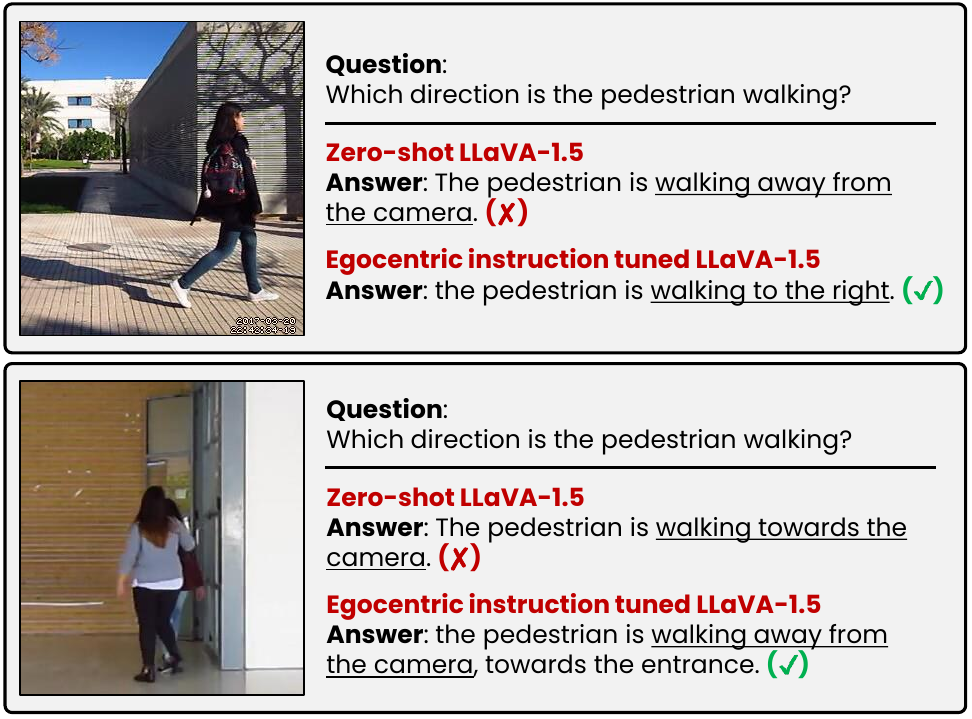}
  \caption{
Pedestrian walking direction prediction results. These examples show that MLLMs with improved object orientation understanding, achieved through our egocentric instruction tuning, can be effectively applied to predicting pedestrian direction.}
  \label{fig:6_pedestrian_direction}
  \vspace{-0.5em}
\end{figure}
\textbf{Pedestrian Walking Direction Prediction.}
If the direction of a pedestrian is accurately determined, their path can be correctly predicted, which helps reduce traffic accidents~\cite{enzweiler2010integrated,dollar2011pedestrian,gandhi2008image}. 
To explore the potential of using MLLMs in pedestrian walking direction prediction, we perform a qualitative analysis using pedestrian walking direction prediction data~\cite{dominguez2017pedestrian}. 

As shown in Figure~\ref{fig:6_pedestrian_direction}, when prompt about a pedestrian’s direction, zero-shot LLaVA-1.5 produces responses that do not match the pedestrian’s actual direction. 
On the other hand, with our egocentric instruction tuning, the model generates responses that accurately align with the correct direction.
Additionally, our method not only accurately describes the orientation of objects (\textit{e.g.}, `walking away from the camera') but also correctly represents the relationship between the object and its surroundings (\textit{e.g.}, `towards the entrance').
Although our egocentric instruction tuning has not been specifically trained on pedestrian direction prediction tasks or images from that domain, it shows improved performance in this task, highlighting its practical utility.

\noindent
\textbf{Spatial Reasoning.}
\begin{table}[t!]
    \centering
    \tiny
    \begin{tabular}{c}
        \includegraphics[width=\columnwidth]{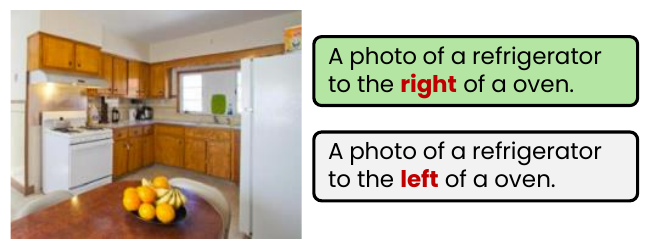} \\ 
    \end{tabular}
    
    \vspace{0.2cm} 

    \resizebox{0.75\columnwidth}{!}{%
    \begin{tabular}{lcc}
        \toprule
        \multirow{2}{*}{Backbone} & \multicolumn{2}{c}{Accuracy}\\
         & Zero-shot & Ours \\
         \midrule
        LLaVA-1.5 &47.5  &\textbf{62.5}  \\
        mPLUG-Owl2 &25.7 &\textbf{40.0} \\
        InternVL &53.9& \textbf{57.3} \\
        \bottomrule
    \end{tabular}%
    }
    \caption{Example and evaluation results for the spatial reasoning task on prepositional relationships between objects. 
    The experimental results demonstrate that MLLMs with enhanced object orientation understanding through Egocentric Instruction Tuning also show improved spatial reasoning abilities for interpreting prepositional relationships between objects.}
    \label{tab:4_spatial_reasoning}
\end{table}

Recently, spatial reasoning research~\cite{comșa2023benchmark,kamath2023s} aimed at understanding spatial relationships between objects using MLLMs has been actively conducted. 
Here, we present that aligning object orientation understanding with the user’s egocentric perspective, thereby enhancing the ability to interpret object orientation itself, can improve MLLM’s spatial reasoning capabilities. 
To support this, we evaluate Zero-shot MLLMs and MLLMs enhanced with our egocentric instruction tuning using the COCO-Spatial evaluation dataset from prior research~\cite{kamath2023s}. 

As shown in the Table~\ref{tab:4_spatial_reasoning}, our egocentric instruction tuning effectively improves spatial reasoning abilities for understanding preposition relationships between objects.
Specifically, accuracy improved by 15.0\%p, 14.3\%p, and 3.4\%p in LLaVA-1.5, mPLUG-Owl2, and InternVL2, respectively.
This indicates that our approach can enhance MLLM performance in user-centered applications that require recognizing object positions, understanding orientations, and interpreting their spatial relationships.

\subsection{Limitations}\label{subsec:6_2_limitation}
\begin{figure}[t]
  \centering
  \includegraphics[width=\columnwidth]{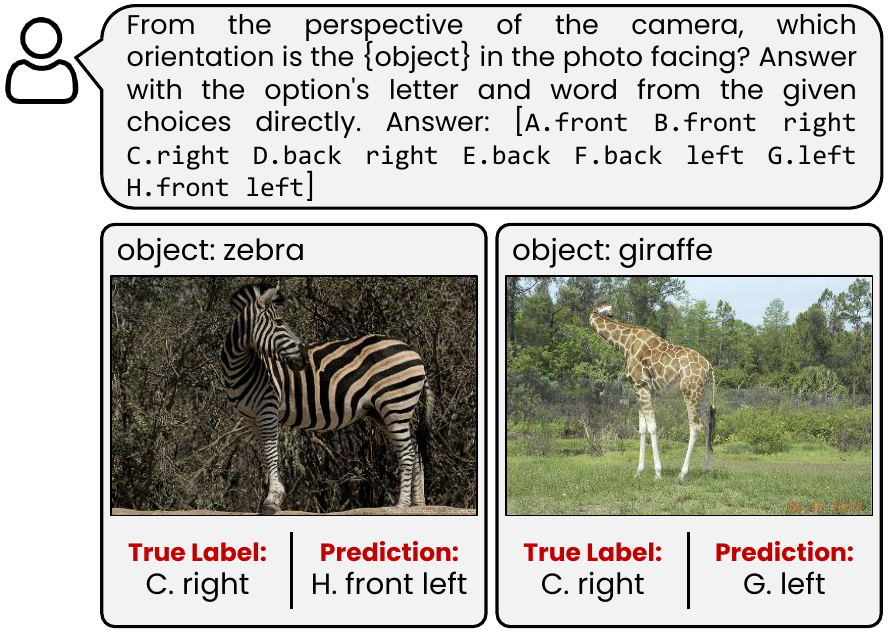}
  \caption{
An example of an error case in our method. Incorrect predictions occur in complex situations where the body and head face different orientations, as seen with the zebra and giraffe.}
  \label{fig:7_limitation}
    \vspace{-0.5em}
\end{figure}
Figure~\ref{fig:7_limitation} presents examples of error cases in LLaVA-1.5 after applying our egocentric instruction tuning.
In these examples, the zebra and giraffe exhibit different orientations for their bodies and heads.
Unlike inanimate objects such as cars or tractors, living beings like zebras and humans can adopt complex, independent orientations.
These complexities pose challenges for our Response Type 2 data, which links object detail recognition with orientation understanding—for instance, when the head and tail suggest different orientations.
Similarly, they complicate Response Type 3 data, which depends on interpreting orientation relationships for alignment tasks.
To address these challenges, future work will explore approaches beyond the current discrete orientation categories.

\section{Conclusion}\label{sec:7_conclusion}
In this study, we identify inconsistent object orientation annotations as a primary factor hindering MLLM's orientation understanding and propose an \textit{Egocentric Instruction Tuning} method to address this by aligning MLLM’s object orientation comprehension with the user’s egocentric perspective. 
Additionally, we introduce \textit{EgoOrientBench}, a benchmark designed to comprehensively evaluate MLLM's orientation understanding across three tasks and five image datasets. 
Experimental results on this benchmark, as well as on the general performance benchmarks, demonstrate that our egocentric instruction tuning effectively enhances object orientation understanding without compromising general response generation capabilities. 
Our analysis also reveals that current MLLMs exhibit a strong bias toward specific orientations, which our instruction tuning mitigates.
Finally, we apply the enhanced MLLM, trained with egocentric instruction tuning, to pedestrian walking direction prediction and spatial reasoning tasks involving spatial relationships between objects. 
This demonstrates the practical benefits of enhanced orientation understanding for real-world applications. 
We believe our instruction tuning method and benchmark will significantly contribute to future research on MLLM's object orientation understanding.
\section*{Acknowledgment}
This work was partly supported by the Institute of Information \& Communications Technology Planning \& Evaluation(IITP)-ICT Creative Consilience Program grant funded by the Korea government(MSIT)(IITP-2025-RS-2020-II201819) and the National Research Foundation of Korea (NRF) grant funded by the Korea government (MSIT) (RS-2023-00212828, RS-2024-00350430).

{
    \small
    \bibliographystyle{ieeenat_fullname}
    \bibliography{main}

\begin{thebibliography}{52}
\providecommand{\natexlab}[1]{#1}
\providecommand{\url}[1]{\texttt{#1}}
\expandafter\ifx\csname urlstyle\endcsname\relax
  \providecommand{\doi}[1]{doi: #1}\else
  \providecommand{\doi}{doi: \begingroup \urlstyle{rm}\Url}\fi

\bibitem[Achiam et~al.(2023)Achiam, Adler, Agarwal, Ahmad, Akkaya, Aleman, Almeida, Altenschmidt, Altman, Anadkat, et~al.]{achiam2023gpt}
Josh Achiam, Steven Adler, Sandhini Agarwal, Lama Ahmad, Ilge Akkaya, Florencia~Leoni Aleman, Diogo Almeida, Janko Altenschmidt, Sam Altman, Shyamal Anadkat, et~al.
\newblock Gpt-4 technical report.
\newblock \emph{arXiv preprint arXiv:2303.08774}, 2023.

\bibitem[Alayrac et~al.(2022)Alayrac, Donahue, Luc, Miech, Barr, Hasson, Lenc, Mensch, Millican, Reynolds, et~al.]{alayrac2022flamingo}
Jean-Baptiste Alayrac, Jeff Donahue, Pauline Luc, Antoine Miech, Iain Barr, Yana Hasson, Karel Lenc, Arthur Mensch, Katherine Millican, Malcolm Reynolds, et~al.
\newblock Flamingo: a visual language model for few-shot learning.
\newblock \emph{Advances in neural information processing systems}, 35:\penalty0 23716--23736, 2022.

\bibitem[Anthropic(2024)]{anthropic2024claude}
AI Anthropic.
\newblock Claude 3.5 sonnet model card addendum.
\newblock \emph{Claude-3.5 Model Card}, 2024.

\bibitem[Bai et~al.(2023)Bai, Bai, Yang, Wang, Tan, Wang, Lin, Zhou, and Zhou]{bai2023qwen}
Jinze Bai, Shuai Bai, Shusheng Yang, Shijie Wang, Sinan Tan, Peng Wang, Junyang Lin, Chang Zhou, and Jingren Zhou.
\newblock Qwen-vl: A frontier large vision-language model with versatile abilities.
\newblock \emph{arXiv preprint arXiv:2308.12966}, 2023.

\bibitem[Chen et~al.(2024{\natexlab{a}})Chen, Xu, Jia, Wang, Liu, Zhang, and Wang]{chen2024egocentric}
Feiyu Chen, Cong Xu, Qi Jia, Yihua Wang, Yuhan Liu, Haotian Zhang, and Endong Wang.
\newblock Egocentric vehicle dense video captioning.
\newblock In \emph{Proceedings of the 32nd ACM International Conference on Multimedia}, pages 137--146, 2024{\natexlab{a}}.

\bibitem[Chen et~al.(2024{\natexlab{b}})Chen, Li, Dong, Zhang, Zang, Chen, Duan, Wang, Qiao, Lin, et~al.]{chen2024we}
Lin Chen, Jinsong Li, Xiaoyi Dong, Pan Zhang, Yuhang Zang, Zehui Chen, Haodong Duan, Jiaqi Wang, Yu Qiao, Dahua Lin, et~al.
\newblock Are we on the right way for evaluating large vision-language models?
\newblock \emph{arXiv preprint arXiv:2403.20330}, 2024{\natexlab{b}}.

\bibitem[Chen et~al.(2024{\natexlab{c}})Chen, Wu, Wang, Su, Chen, Xing, Zhong, Zhang, Zhu, Lu, et~al.]{chen2024internvl}
Zhe Chen, Jiannan Wu, Wenhai Wang, Weijie Su, Guo Chen, Sen Xing, Muyan Zhong, Qinglong Zhang, Xizhou Zhu, Lewei Lu, et~al.
\newblock Internvl: Scaling up vision foundation models and aligning for generic visual-linguistic tasks.
\newblock In \emph{Proceedings of the IEEE/CVF Conference on Computer Vision and Pattern Recognition}, pages 24185--24198, 2024{\natexlab{c}}.

\bibitem[Comșa and Narayanan(2023)]{comșa2023benchmark}
Iulia Comșa and Srini Narayanan.
\newblock A benchmark for reasoning with spatial prepositions.
\newblock In \emph{Proceedings of the 2023 Conference on Empirical Methods in Natural Language Processing}, pages 16328--16335, 2023.

\bibitem[Dai et~al.(2023)Dai, Li, Li, Tiong, Zhao, Wang, Li, Fung, and Hoi]{dai2023instructblipgeneralpurposevisionlanguagemodels}
Wenliang Dai, Junnan Li, Dongxu Li, Anthony Meng~Huat Tiong, Junqi Zhao, Weisheng Wang, Boyang Li, Pascale Fung, and Steven Hoi.
\newblock Instructblip: Towards general-purpose vision-language models with instruction tuning, 2023.

\bibitem[Deng et~al.(2009)Deng, Dong, Socher, Li, Li, and Fei-Fei]{deng2009imagenet}
Jia Deng, Wei Dong, Richard Socher, Li-Jia Li, Kai Li, and Li Fei-Fei.
\newblock Imagenet: A large-scale hierarchical image database.
\newblock In \emph{2009 IEEE conference on computer vision and pattern recognition}, pages 248--255. Ieee, 2009.

\bibitem[Di et~al.(2022)Di, Zhang, Lou, Manhardt, Ji, Navab, and Tombari]{di2022gpv}
Yan Di, Ruida Zhang, Zhiqiang Lou, Fabian Manhardt, Xiangyang Ji, Nassir Navab, and Federico Tombari.
\newblock Gpv-pose: Category-level object pose estimation via geometry-guided point-wise voting.
\newblock In \emph{Proceedings of the IEEE/CVF Conference on Computer Vision and Pattern Recognition}, pages 6781--6791, 2022.

\bibitem[Dollar et~al.(2011)Dollar, Wojek, Schiele, and Perona]{dollar2011pedestrian}
Piotr Dollar, Christian Wojek, Bernt Schiele, and Pietro Perona.
\newblock Pedestrian detection: An evaluation of the state of the art.
\newblock \emph{IEEE transactions on pattern analysis and machine intelligence}, 34\penalty0 (4):\penalty0 743--761, 2011.

\bibitem[Dominguez-Sanchez et~al.(2017)Dominguez-Sanchez, Cazorla, and Orts-Escolano]{dominguez2017pedestrian}
Alex Dominguez-Sanchez, Miguel Cazorla, and Sergio Orts-Escolano.
\newblock Pedestrian movement direction recognition using convolutional neural networks.
\newblock \emph{IEEE transactions on intelligent transportation systems}, 18\penalty0 (12):\penalty0 3540--3548, 2017.

\bibitem[Driess et~al.(2023)Driess, Xia, Sajjadi, Lynch, Chowdhery, Ichter, Wahid, Tompson, Vuong, Yu, et~al.]{driess2023palm}
Danny Driess, Fei Xia, Mehdi~SM Sajjadi, Corey Lynch, Aakanksha Chowdhery, Brian Ichter, Ayzaan Wahid, Jonathan Tompson, Quan Vuong, Tianhe Yu, et~al.
\newblock Palm-e: An embodied multimodal language model.
\newblock In \emph{International Conference on Machine Learning}, pages 8469--8488. PMLR, 2023.

\bibitem[Enzweiler and Gavrila(2010)]{enzweiler2010integrated}
Markus Enzweiler and Dariu~M Gavrila.
\newblock Integrated pedestrian classification and orientation estimation.
\newblock In \emph{2010 IEEE Computer Society Conference on Computer Vision and Pattern Recognition}, pages 982--989. IEEE, 2010.

\bibitem[Fu et~al.(2023)Fu, Chen, Shen, Qin, Zhang, Lin, Yang, Zheng, Li, Sun, et~al.]{fu2023mme}
Chaoyou Fu, Peixian Chen, Yunhang Shen, Yulei Qin, Mengdan Zhang, Xu Lin, Jinrui Yang, Xiawu Zheng, Ke Li, Xing Sun, et~al.
\newblock Mme: A comprehensive evaluation benchmark for multimodal large language models.
\newblock \emph{arXiv preprint arXiv:2306.13394}, 2023.

\bibitem[Gandhi and Trivedi(2008)]{gandhi2008image}
Tarak Gandhi and Mohan~Manubhai Trivedi.
\newblock Image based estimation of pedestrian orientation for improving path prediction.
\newblock In \emph{2008 IEEE Intelligent Vehicles Symposium}, pages 506--511. IEEE, 2008.

\bibitem[Gao et~al.(2024)Gao, Sarkar, Xia, Xiao, Wu, Ichter, Majumdar, and Sadigh]{gao2023physically}
Jensen Gao, Bidipta Sarkar, Fei Xia, Ted Xiao, Jiajun Wu, Brian Ichter, Anirudha Majumdar, and Dorsa Sadigh.
\newblock Physically grounded vision-language models for robotic manipulation.
\newblock In \emph{International Conference on Robotics and Automation (ICRA)}, 2024.

\bibitem[Gaur et~al.(2024)Gaur, Tapaswi, et~al.]{gaur2024detect}
Manu Gaur, Makarand Tapaswi, et~al.
\newblock Detect, describe, discriminate: Moving beyond vqa for mllm evaluation.
\newblock \emph{arXiv preprint arXiv:2409.15125}, 2024.

\bibitem[Gramann et~al.(2010)Gramann, Onton, Riccobon, Mueller, Bardins, and Makeig]{gramann2010human}
Klaus Gramann, Julie Onton, Davide Riccobon, Hermann~J Mueller, Stanislav Bardins, and Scott Makeig.
\newblock Human brain dynamics accompanying use of egocentric and allocentric reference frames during navigation.
\newblock \emph{Journal of cognitive neuroscience}, 22\penalty0 (12):\penalty0 2836--2849, 2010.

\bibitem[Grauman et~al.(2022)Grauman, Westbury, Byrne, Chavis, Furnari, Girdhar, Hamburger, Jiang, Liu, Liu, et~al.]{grauman2022ego4d}
Kristen Grauman, Andrew Westbury, Eugene Byrne, Zachary Chavis, Antonino Furnari, Rohit Girdhar, Jackson Hamburger, Hao Jiang, Miao Liu, Xingyu Liu, et~al.
\newblock Ego4d: Around the world in 3,000 hours of egocentric video.
\newblock In \emph{Proceedings of the IEEE/CVF Conference on Computer Vision and Pattern Recognition}, pages 18995--19012, 2022.

\bibitem[Guan et~al.(2024)Guan, Yang, Cheng, Lin, Kim, Madhivanan, Sen, and Manocha]{guan2024loc}
Tianrui Guan, Yurou Yang, Harry Cheng, Muyuan Lin, Richard Kim, Rajasimman Madhivanan, Arnie Sen, and Dinesh Manocha.
\newblock Loc-zson: Language-driven object-centric zero-shot object retrieval and navigation.
\newblock In \emph{International Conference on Robotics and Automation (ICRA)}, 2024.

\bibitem[Hu et~al.()Hu, Wallis, Allen-Zhu, Li, Wang, Wang, Chen, et~al.]{hulora}
Edward~J Hu, Phillip Wallis, Zeyuan Allen-Zhu, Yuanzhi Li, Shean Wang, Lu Wang, Weizhu Chen, et~al.
\newblock Lora: Low-rank adaptation of large language models.
\newblock In \emph{International Conference on Learning Representations}.

\bibitem[Huang et~al.(2024)Huang, Chen, Xu, Zhang, Yang, Pei, Zhang, Dong, Wang, Wang, et~al.]{huang2024egoexolearn}
Yifei Huang, Guo Chen, Jilan Xu, Mingfang Zhang, Lijin Yang, Baoqi Pei, Hongjie Zhang, Lu Dong, Yali Wang, Limin Wang, et~al.
\newblock Egoexolearn: A dataset for bridging asynchronous ego-and exo-centric view of procedural activities in real world.
\newblock In \emph{Proceedings of the IEEE/CVF Conference on Computer Vision and Pattern Recognition}, pages 22072--22086, 2024.

\bibitem[Kamath et~al.(2023)Kamath, Hessel, and Chang]{kamath2023s}
Amita Kamath, Jack Hessel, and Kai-Wei Chang.
\newblock What’s “up” with vision-language models? investigating their struggle with spatial reasoning.
\newblock In \emph{Proceedings of the 2023 Conference on Empirical Methods in Natural Language Processing}, pages 9161--9175, 2023.

\bibitem[Levinson(1996)]{levinson1996frames}
Stephen~C Levinson.
\newblock Frames of reference and molyneux's question: Crosslinguistic evidence.
\newblock 1996.

\bibitem[Li et~al.(2017)Li, Yang, Song, and Hospedales]{li2017deeper}
Da Li, Yongxin Yang, Yi-Zhe Song, and Timothy~M Hospedales.
\newblock Deeper, broader and artier domain generalization.
\newblock In \emph{Proceedings of the IEEE international conference on computer vision}, pages 5542--5550, 2017.

\bibitem[Li et~al.(2023)Li, Du, Zhou, Wang, Zhao, and Wen]{li2023evaluating}
Yifan Li, Yifan Du, Kun Zhou, Jinpeng Wang, Wayne~Xin Zhao, and Ji-Rong Wen.
\newblock Evaluating object hallucination in large vision-language models.
\newblock \emph{arXiv preprint arXiv:2305.10355}, 2023.

\bibitem[Lin et~al.(2014)Lin, Maire, Belongie, Hays, Perona, Ramanan, Doll{\'a}r, and Zitnick]{lin2014microsoft}
Tsung-Yi Lin, Michael Maire, Serge Belongie, James Hays, Pietro Perona, Deva Ramanan, Piotr Doll{\'a}r, and C~Lawrence Zitnick.
\newblock Microsoft coco: Common objects in context.
\newblock In \emph{Computer Vision--ECCV 2014: 13th European Conference, Zurich, Switzerland, September 6-12, 2014, Proceedings, Part V 13}, pages 740--755. Springer, 2014.

\bibitem[Liu et~al.(2024)Liu, Li, Wu, and Lee]{liu2024visual}
Haotian Liu, Chunyuan Li, Qingyang Wu, and Yong~Jae Lee.
\newblock Visual instruction tuning.
\newblock \emph{Advances in neural information processing systems}, 36, 2024.

\bibitem[Loshchilov and Hutter(2019)]{loshchilov2018decoupled}
Ilya Loshchilov and Frank Hutter.
\newblock Decoupled weight decay regularization.
\newblock In \emph{International Conference on Learning Representations}, 2019.

\bibitem[Mangalam et~al.(2023)Mangalam, Akshulakov, and Malik]{mangalam2023egoschema}
Karttikeya Mangalam, Raiymbek Akshulakov, and Jitendra Malik.
\newblock Egoschema: A diagnostic benchmark for very long-form video language understanding.
\newblock \emph{Advances in Neural Information Processing Systems}, 36:\penalty0 46212--46244, 2023.

\bibitem[Mu et~al.(2024)Mu, Zhang, Hu, Wang, Ding, Jin, Wang, Dai, Qiao, and Luo]{mu2024embodiedgpt}
Yao Mu, Qinglong Zhang, Mengkang Hu, Wenhai Wang, Mingyu Ding, Jun Jin, Bin Wang, Jifeng Dai, Yu Qiao, and Ping Luo.
\newblock Embodiedgpt: Vision-language pre-training via embodied chain of thought.
\newblock \emph{Advances in Neural Information Processing Systems}, 36, 2024.

\bibitem[Ozuysal et~al.(2009)Ozuysal, Lepetit, and Fua]{ozuysal2009pose}
Mustafa Ozuysal, Vincent Lepetit, and Pascal Fua.
\newblock Pose estimation for category specific multiview object localization.
\newblock In \emph{2009 IEEE Conference on Computer Vision and Pattern Recognition}, pages 778--785. IEEE, 2009.

\bibitem[Pei et~al.(2024)Pei, Viola, Huang, Wang, Ahsan, Ye, Yiming, Sai, Wang, Chen, et~al.]{pei2024autonomous}
Jiahuan Pei, Irene Viola, Haochen Huang, Junxiao Wang, Moonisa Ahsan, Fanghua Ye, Jiang Yiming, Yao Sai, Di Wang, Zhumin Chen, et~al.
\newblock Autonomous workflow for multimodal fine-grained training assistants towards mixed reality.
\newblock \emph{arXiv preprint arXiv:2405.13034}, 2024.

\bibitem[Peng et~al.(2019)Peng, Bai, Xia, Huang, Saenko, and Wang]{peng2019moment}
Xingchao Peng, Qinxun Bai, Xide Xia, Zijun Huang, Kate Saenko, and Bo Wang.
\newblock Moment matching for multi-source domain adaptation.
\newblock In \emph{Proceedings of the IEEE/CVF international conference on computer vision}, pages 1406--1415, 2019.

\bibitem[Peng et~al.(2023)Peng, Wang, Dong, Hao, Huang, Ma, and Wei]{peng2023kosmos}
Zhiliang Peng, Wenhui Wang, Li Dong, Yaru Hao, Shaohan Huang, Shuming Ma, and Furu Wei.
\newblock Kosmos-2: Grounding multimodal large language models to the world.
\newblock \emph{arXiv preprint arXiv:2306.14824}, 2023.

\bibitem[Poole et~al.()Poole, Jain, Barron, and Mildenhall]{pooledreamfusion}
Ben Poole, Ajay Jain, Jonathan~T Barron, and Ben Mildenhall.
\newblock Dreamfusion: Text-to-3d using 2d diffusion.
\newblock In \emph{The Eleventh International Conference on Learning Representations}.

\bibitem[Schuhmann et~al.(2022)Schuhmann, Beaumont, Vencu, Gordon, Wightman, Cherti, Coombes, Katta, Mullis, Wortsman, et~al.]{schuhmann2022laion}
Christoph Schuhmann, Romain Beaumont, Richard Vencu, Cade Gordon, Ross Wightman, Mehdi Cherti, Theo Coombes, Aarush Katta, Clayton Mullis, Mitchell Wortsman, et~al.
\newblock Laion-5b: An open large-scale dataset for training next generation image-text models.
\newblock \emph{Advances in Neural Information Processing Systems}, 35:\penalty0 25278--25294, 2022.

\bibitem[Shao et~al.(2024)Shao, Hu, Wang, Song, Waslander, Liu, and Li]{shao2024lmdrive}
Hao Shao, Yuxuan Hu, Letian Wang, Guanglu Song, Steven~L Waslander, Yu Liu, and Hongsheng Li.
\newblock Lmdrive: Closed-loop end-to-end driving with large language models.
\newblock In \emph{Proceedings of the IEEE/CVF Conference on Computer Vision and Pattern Recognition}, pages 15120--15130, 2024.

\bibitem[Team()]{team5gemini}
G~Gemini Team.
\newblock Gemini 1.5: Unlocking multimodal understanding across millions of tokens of context (2024).
\newblock \emph{URL https://goo. gle/GeminiV1-5}.

\bibitem[Tong et~al.(2024)Tong, Liu, Zhai, Ma, LeCun, and Xie]{tong2024eyes}
Shengbang Tong, Zhuang Liu, Yuexiang Zhai, Yi Ma, Yann LeCun, and Saining Xie.
\newblock Eyes wide shut? exploring the visual shortcomings of multimodal llms.
\newblock In \emph{Proceedings of the IEEE/CVF Conference on Computer Vision and Pattern Recognition}, pages 9568--9578, 2024.

\bibitem[Vallar et~al.(1999)Vallar, Lobel, Galati, Berthoz, Pizzamiglio, and Le~Bihan]{vallar1999fronto}
Giuseppe Vallar, Elie Lobel, Gaspare Galati, Alain Berthoz, Luigi Pizzamiglio, and Denis Le~Bihan.
\newblock A fronto-parietal system for computing the egocentric spatial frame of reference in humans.
\newblock \emph{Experimental brain research}, 124:\penalty0 281--286, 1999.

\bibitem[Wang et~al.(2024)Wang, Chen, Chen, Wu, Zhu, Zeng, Luo, Lu, Zhou, Qiao, et~al.]{wang2024visionllm}
Wenhai Wang, Zhe Chen, Xiaokang Chen, Jiannan Wu, Xizhou Zhu, Gang Zeng, Ping Luo, Tong Lu, Jie Zhou, Yu Qiao, et~al.
\newblock Visionllm: Large language model is also an open-ended decoder for vision-centric tasks.
\newblock \emph{Advances in Neural Information Processing Systems}, 36, 2024.

\bibitem[Wang et~al.(2023)Wang, Kwon, Rad, Pan, Chakraborty, Andrist, Bohus, Feniello, Tekin, Frujeri, et~al.]{wang2023holoassist}
Xin Wang, Taein Kwon, Mahdi Rad, Bowen Pan, Ishani Chakraborty, Sean Andrist, Dan Bohus, Ashley Feniello, Bugra Tekin, Felipe~Vieira Frujeri, et~al.
\newblock Holoassist: an egocentric human interaction dataset for interactive ai assistants in the real world.
\newblock In \emph{Proceedings of the IEEE/CVF International Conference on Computer Vision}, pages 20270--20281, 2023.

\bibitem[Wei et~al.(2024)Wei, Wang, Lu, Xu, Liu, Zhao, Chen, and Wang]{wei2024editable}
Yuxi Wei, Zi Wang, Yifan Lu, Chenxin Xu, Changxing Liu, Hao Zhao, Siheng Chen, and Yanfeng Wang.
\newblock Editable scene simulation for autonomous driving via collaborative llm-agents.
\newblock In \emph{Proceedings of the IEEE/CVF Conference on Computer Vision and Pattern Recognition}, pages 15077--15087, 2024.

\bibitem[Wu et~al.(2023)Wu, Zhang, Fu, Wang, Ren, Pan, Wu, Yang, Wang, Qian, et~al.]{wu2023omniobject3d}
Tong Wu, Jiarui Zhang, Xiao Fu, Yuxin Wang, Jiawei Ren, Liang Pan, Wayne Wu, Lei Yang, Jiaqi Wang, Chen Qian, et~al.
\newblock Omniobject3d: Large-vocabulary 3d object dataset for realistic perception, reconstruction and generation.
\newblock In \emph{Proceedings of the IEEE/CVF Conference on Computer Vision and Pattern Recognition}, pages 803--814, 2023.

\bibitem[Xu et~al.(2024)Xu, Huang, Hou, Chen, Zhang, Feng, and Xie]{xu2024retrieval}
Jilan Xu, Yifei Huang, Junlin Hou, Guo Chen, Yuejie Zhang, Rui Feng, and Weidi Xie.
\newblock Retrieval-augmented egocentric video captioning.
\newblock In \emph{Proceedings of the IEEE/CVF Conference on Computer Vision and Pattern Recognition}, pages 13525--13536, 2024.

\bibitem[Ye et~al.(2024)Ye, Xu, Ye, Yan, Hu, Liu, Qian, Zhang, and Huang]{ye2024mplug}
Qinghao Ye, Haiyang Xu, Jiabo Ye, Ming Yan, Anwen Hu, Haowei Liu, Qi Qian, Ji Zhang, and Fei Huang.
\newblock mplug-owl2: Revolutionizing multi-modal large language model with modality collaboration.
\newblock In \emph{Proceedings of the IEEE/CVF Conference on Computer Vision and Pattern Recognition}, pages 13040--13051, 2024.

\bibitem[Yue et~al.(2024)Yue, Ni, Zhang, Zheng, Liu, Zhang, Stevens, Jiang, Ren, Sun, et~al.]{yue2024mmmu}
Xiang Yue, Yuansheng Ni, Kai Zhang, Tianyu Zheng, Ruoqi Liu, Ge Zhang, Samuel Stevens, Dongfu Jiang, Weiming Ren, Yuxuan Sun, et~al.
\newblock Mmmu: A massive multi-discipline multimodal understanding and reasoning benchmark for expert agi.
\newblock In \emph{Proceedings of the IEEE/CVF Conference on Computer Vision and Pattern Recognition}, pages 9556--9567, 2024.

\bibitem[Zaehle et~al.(2007)Zaehle, Jordan, W{\"u}stenberg, Baudewig, Dechent, and Mast]{zaehle2007neural}
Tino Zaehle, Kirsten Jordan, Torsten W{\"u}stenberg, J{\"u}rgen Baudewig, Peter Dechent, and Fred~W Mast.
\newblock The neural basis of the egocentric and allocentric spatial frame of reference.
\newblock \emph{Brain research}, 1137:\penalty0 92--103, 2007.

\bibitem[Zhu et~al.(2024)Zhu, Chen, Shen, Li, and Elhoseiny]{zhu2024minigpt}
Deyao Zhu, Jun Chen, Xiaoqian Shen, Xiang Li, and Mohamed Elhoseiny.
\newblock Minigpt-4: Enhancing vision-language understanding with advanced large language models.
\newblock In \emph{The Twelfth International Conference on Learning Representations}, 2024.

\end{thebibliography}
}

\clearpage
\appendix
\section*{Appendix}
\pagenumbering{arabic}
\renewcommand*{\thepage}{A\arabic{page}}
\section{Data Collection}\label{sec:a_data_collection}

\subsection{Instruction Data Generation}\label{subsec:a_1_instruction_data_generation}
We generate three types of data using the inference capabilities of the LLaVA1.5 Vicuna 13B model~\cite{liu2024visual}. 
The model is prompted to produce questions and answers based on a given context, and the outputs are parsed to construct instruction learning data. 
The prompts corresponding to each data type are described in Table~\ref{tab:template_sample}.

\subsection{Data Statistics}\label{subsec:a_2_data_statistics}
\textbf{Training Dataset.}
We collect data from the ImageNet dataset~\cite{deng2009imagenet}.
We then remove ambiguous images for determining orientations, resulting in a dataset of 2,845 images. 
To ensure a uniform number of evaluation data per orientation class, we separate benchmark data from the source dataset by extracting 50 images per orientation. 
The remaining 2,445 images are used to construct the training dataset. 
For each image, we generate three questions corresponding to the three types of data described earlier. 
As a result, the training dataset consists of a total of 2,445 images, with three instruction data points generated per image, leading to a total of 7,335 data points. 
The statistics of the training data is shown in Table ~\ref{tab:5_Egocentric_Instruction_Data}.
\begin{table}[!h]
\centering
\small
\begin{tabular}{ll}
\toprule
\textbf{Class}       & \textbf{Count(IN)} \\ \midrule
\textit{Front-Left}  & 1,284                      \\
\textit{Front}       & 618                       \\
\textit{Front-Right} & 1,326                      \\
\textit{Right}       & 1,689                      \\
\textit{Back-Right}  & 225                       \\
\textit{Back}        & 84                        \\
\textit{Back-Left}   & 270                       \\
\textit{Left}        & 1,839                      \\ 
\bottomrule
\end{tabular}
\caption{The number of data samples for each orientation class.}
\label{tab:5_Egocentric_Instruction_Data}
\end{table}

\noindent
\textbf{Benchmark Data.}
Table~\ref{tab:6_benchmark_statistics} presents the statistics of orientation classes for the images in our benchmark. 
The ImageNet~\cite{deng2009imagenet} and OmniObject3D~\cite{wu2023omniobject3d} datasets are controlled to exhibit a uniform distribution, while the remaining datasets show an imbalanced distribution. 
Each image is used once for the \textit{Choose} and \textit{Freeform} tasks and twice for the \textit{Verify} task, resulting in the distribution of data samples for each task, as shown in Table~\ref{tab:7_benchmark_statistics}.

\begin{table}[!t]
\centering
\small

\begin{tabular}{llllll}
\toprule
\multicolumn{1}{l}{\multirow{2}{*}{Class}}&\multicolumn{5}{c}{Images}\\
& IN & D$_3$ & DN & PA & 3D  \\\midrule

\textit{Front-Left} & 50 & 14 & 197 & 171 & 500 \\
\textit{Front} & 50 & 38 & 422 & 517 & 500 \\
\textit{Front-Right} & 50 & 29 & 213 & 156 & 500\\
\textit{Right} & 50 & 29 & 367 & 586 & 500 \\
\textit{Back-Right} & 50 & 4 & 14 & 10 & 500 \\
\textit{Back} & 50 & 2 & 18 & 3 & 500 \\
\textit{Back-Left} & 50 & 4 & 17 & 5 & 500 \\
\textit{Left} & 50 & 32 & 477 & 640 & 500 \\
\bottomrule
\end{tabular}
\caption{Class-wise data statistics of collected images in our benchmark. Our benchmark is constructed with manually collected orientation annotations from five datasets, ImageNet (IN)~\cite{deng2009imagenet}, D$_3$~\cite{gaur2024detect}, DomainNet (DN)~\cite{peng2019moment}, PACS (PA)~\cite{li2017deeper}, and OmniObject3D (3D)~\cite{wu2023omniobject3d}.}
\label{tab:6_benchmark_statistics}
\end{table}

\begin{table}[!t]
\centering
\small

\begin{tabular}{llllll}
\toprule
\multicolumn{1}{l}{\multirow{2}{*}{Task}}&\multicolumn{5}{c}{Datasets}\\
& IN & D$_3$ & DN & PA & 3D  \\\midrule

\textit{Choose} & 400 & 152 & 1,725 & 2,088 & 4,000 \\
\textit{Verify} & 800 & 304 & 3,450 & 4,176 & 8,000 \\
\textit{Freeform} & 400 & 152 & 1,725 & 2,088 & 4,000 \\
\bottomrule
\end{tabular}
\caption{Task-wise data statistics of our benchmark. Our benchmark is constructed with manually collected orientation annotations from five datasets, ImageNet (IN)~\cite{deng2009imagenet}, D$_3$~\cite{gaur2024detect}, DomainNet (DN)~\cite{peng2019moment}, PACS (PA)~\cite{li2017deeper}, and OmniObject3D (3D)~\cite{wu2023omniobject3d}.}
\label{tab:7_benchmark_statistics}
\end{table}

\subsection{Task Data Details}\label{subsec:a_3_task_data_details}
\begin{itemize}
    \item \textit{Choose}: We prompt the model to select the direction an object in the image is facing from among eight directional options. This task is designed to evaluate the basic capability of MLLMs to recognize general directions.
    
    \item \textit{Verify}: In this task, there may be a bias towards ``yes” or ``no” answers, which could result in evaluation metrics failing to accurately represent the actual performance. To address this issue, we design two separate verification tasks for each image: one where the correct answer is ``yes” and another where the expected answer is ``no.”
    Specifically, for a given image, if the orientation of the image is labeled as ``right," we first construct a question asking whether the subject is facing ``right." 
    Then, we randomly select one of the remaining orientations (excluding ``right" from the eight possible orientations) and create a question asking whether the subject is facing the selected orientation.
    
    \item \textit{Freeform}: We recognize that expressions for orientation can vary significantly. For example, phrases like ``toward the front,” ``facing forward,” or ``looking ahead” may all describe the same orientation. To account for these variations, we allow free-form responses and include a ``freeform" metric, verified using the GPT-4o API (\texttt{gpt-4o-2024-08-06}), as an additional evaluation measure.

\end{itemize}
The data format for each task is summarized in Table~\ref{tab:8_benchmark_samples}.

\begin{table}[!t]
    \centering
    \normalsize
    \resizebox{\columnwidth}{!}{
    \begin{tabular}{p{1.5cm}p{7cm}} 
        \toprule
        \textbf{Class} & \textbf{Example} \\ 
        \midrule
        \textit{Choose} & 
        From the perspective of the camera, which orientation is the \{Object\} in the photo facing? A. front B. front right C. right D. back right E. back F. back left G. left H. front left. Answer with the option's letter and word from the given choices directly. \\ \midrule
        
        \textit{Verify} & 
        Is the \{Object\} facing ``front right" from the camera's perspective? Answer with ``yes" or ``no" only. \\ \midrule
        
        \textit{Freeform} & 
        From the perspective of the camera, Answer what orientation the \{Object\} in the picture is facing. \\
        \bottomrule
    \end{tabular}%
    }
    \caption{The data format for each task.}
    \label{tab:8_benchmark_samples}
\end{table}

\subsection{3D Rendered Images}\label{subsec:a_4_3d_rendered_images}
We utilize the OmniObject3D dataset~\cite{wu2023omniobject3d}, which includes high-quality, real-scanned 3D objects designed to advance 3D perception, reconstruction, and generation tasks in real-world scenarios. 
We use a total of 500 3D scans distributed across 11 object categories.
Each scan is oriented along a specific axis ($+x, -x, +y, -y, +z, -z$), and we manually unify the coordinate system to standardize their orientations. 
Subsequently, we place eight cameras (Front-Left, Front, Front-Right, Left, Right, Back-Left, Back, Back-Right) around the object at 45° intervals with respect to the object's center. 
For the front and back views, the cameras are tilted downward by approximately 20° to better capture the object's orientation. 
Through this process, we render a total of 4,000 RGB images with a 448×448 image resolution. 

\section{Experimental Details}
We use GPT-4o to evaluate the models on the \textit{Freeform} task and spatial reasoning. 
The prompts used for GPT-4o are detailed in Table~\ref{tab:9_gpt_prompts}.

\begin{table}[!t]
    \centering
    \normalsize
    \resizebox{\columnwidth}{!}{%
    \begin{tabular}{p{1.8cm}p{8cm}} 
        \toprule
        \textbf{Class} & \textbf{Example} \\ 
        \midrule

        \textit{GPT Evaluation} & 
        \begin{minipage}[t]{8cm} 
        You are given an answer and a prediction representing an object’s orientation out of 8 possible directions. 
        Respond with `yes' if the answer and prediction match, or `no' if they do not.

        \vspace{0.2cm}
        [Example]
        If the answer is `front right' and the prediction is `facing right while facing the camera,' respond with `yes.'
        If the answer is `front right' and the prediction is `facing the camera,' respond with `no,' because `front' and `front right' differ in orientation.
        
        \vspace{0.2cm}
        Answer: \{answer\} Prediction:\{prediction\}
        \end{minipage} \\ \midrule

        \textit{Preposition Prompt} &  
        \begin{minipage}[t]{8cm} 
           From the perspective of the camera,
           look at the given photo and choose the sentence that best describes 
           its content between the two options. 
           
           \vspace{0.2cm}
           A. \{option a\} B. \{option b\}
        \end{minipage} \\ \midrule

        \textit{Preposition GPT Eval} &  
        \begin{minipage}[t]{8cm} 
            \raggedright
            Check if the given prediction matches the ground truth. 
            Respond with `yes' only if they match, and `no' otherwise.
            
            \vspace{0.2cm}
            Letter Answer: A. Sentence Answer: \{answer\}, Prediction: \{prediction\}
        \end{minipage} \\ 

       \bottomrule
    \end{tabular}%
    }
    \caption{Table with detailed row descriptions}
    \label{tab:9_gpt_prompts}
\end{table}

\section{Further Analysis}\label{sec:c_further_analysis}
\subsection{Confusion Matrix}\label{subsec:c_1_confusion_matrix}
\begin{figure}[h]
  \centering
  \includegraphics[width=\columnwidth]{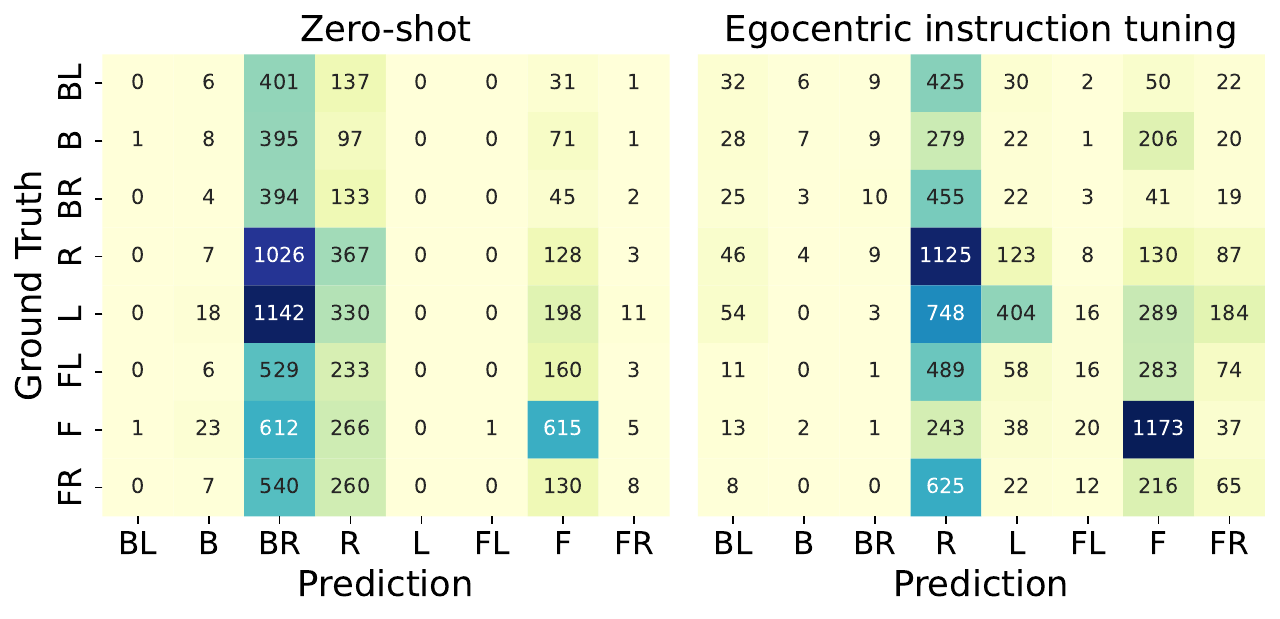}
  \caption{Confusion matrix for the \textit{Choose} task with InternVL.}
  \label{fig:8_confusion_matrix_internVL.pdf}
\end{figure}

We draw confusion matrix for InternVL~\cite{chen2024internvl} in Figure~\ref{fig:8_confusion_matrix_internVL.pdf}.
As with the confusion matrices of other models, it can be observed that the responses of the model become more aligned after training, and the biases are mitigated.

\subsection{Ablation Test}\label{subsec:c_2_ablation_test}
Table~\ref{tab:10_more_ablation_test} shows the additional results of ablation tests for mPLUG-Owl2~\cite{ye2024mplug} and InternVL~\cite{chen2024internvl}.
Although not all data types show the same upward trend across every metric for all models, one consistent observation is that the highest performance is achieved when all three response types are utilized. 
In all tested models, removing any response type resulted in a decrease in task accuracy, confirming that each type contributes to improving the MLLM's understanding of object orientation.

\begin{table}[!t]
\centering
\scriptsize
\resizebox{\columnwidth}{!}{%
\begin{tabular}{cccccc}
\toprule
\multicolumn{3}{c}{Response Types} & \multirow{2}{*}{Choose} & \multirow{2}{*}{Verify} & \multirow{2}{*}{Freeform} \\
Type 1 & Type 2 & Type 3 & & & \\\midrule
\cmark & \xmark & \xmark & 25.5 & 51.7 & 30.7 \\
\xmark & \cmark & \xmark & 18.8 & 58.3 & 28.6 \\
\xmark & \xmark & \cmark & 21.2 & 58.9 & 29.1 \\
\midrule
\cmark & \cmark & \xmark & 23.9 & 59.3 & 33.8 \\
\cmark & \xmark & \cmark & 28.4 & 59.9 & 33.4 \\
\xmark & \cmark & \cmark & 21.0 & 58.1 & 34.2 \\
\midrule
\cmark & \cmark & \cmark & 28.5 & 61.5 & 37.1 \\
\bottomrule
\end{tabular}%
}

\vspace{0.2cm} 

\resizebox{\columnwidth}{!}{%
\begin{tabular}{cccccc}
\toprule
\multicolumn{3}{c}{Response Types} & \multirow{2}{*}{Choose} & \multirow{2}{*}{Verify} & \multirow{2}{*}{Freeform} \\
Type 1 & Type 2 & Type 3 & & & \\\midrule
\cmark & \xmark & \xmark & 17.4 & 56.8 & 18.9 \\
\xmark & \cmark & \xmark & 17.9 & 54.2 & 32.7 \\
\xmark & \xmark & \cmark & 20.7 & 58.1 & 41.6 \\
\midrule
\cmark & \cmark & \xmark & 22.1 & 58.2 & 31.1 \\
\cmark & \xmark & \cmark & 26.5 & 61.2 & 43.6 \\
\xmark & \cmark & \cmark & 24.3 & 56.1 & 44.1 \\
\midrule
\cmark & \cmark & \cmark & 31.4 & 61.4 & 48.2 \\
\bottomrule
\end{tabular}%
}
\caption{Ablation test results for both mPLUG-Owl2 (top) and internVL (bottom). Each response type contributes to performance improvements across all tasks.}
\label{tab:10_more_ablation_test}
\end{table}

\begin{table*}[t]
\centering
\small
\resizebox{\textwidth}{!}{%
\begin{tabular}{p{3.5cm}|p{14.5cm}} 
\toprule

\textbf{Data Type1 Prompt} &
As a competent assistant, your role is to explain the subparts of the main object and, based on your findings, determine its orientation.
    
These parts of the object should eventually be able to imply some orientation of the object, but questions should never directly include information about its orientation. 

You must use the information in [Context], but the important thing is that you must find subparts or sub-features of the object given in [Context].

If you identify the main object as a human, you need to find different subparts depending on the orientation the person is facing in the picture.
For example, if the person is facing forward (i.e., looking at the camera), you will see the face, eyes, chest, or abdomen.

If the person is facing backward, you will see the hip, back, or hair. If the person is facing to the right, you will see one ear and one arm, and you must note that the nose is pointing to the right.

For example, 
 [Question]: From camera perspective, does the \{sport car\} is \{facing camera\} or \{facing away\} the camera/observer? First describe what you can find from the object in the image, then based on that, answer the orientation of the object. 
 [Answer]: (What I find): headlight, windshield, bumper (Answer the orientation): According to (What I find), The \{sport car\} facing the camera/observe.
 
 [Question]: Does the \{a girl\} is facing left or facing right from camera perspective? First describe what you can find from the object in the image, then based on that, answer the orientation of the object.  
 [Answer]: (finding): hair, hip, arm, half of nose (Answer the orientation): According to (What I find), The \{a girl\} facing the left.

    \vspace{0.2cm}
    Now your turn:
    +  context + ``[Question]: " + ``[Answer]:  " \\ \midrule

\textbf{Data Type2 Prompt} &
As a capable vision-language model assistant, your task is to closely examine key features of the object in the provided image and perform the following actions: 
1) identify and ask questions about the features that indicate the object's orientation, and 2) answer the question yourself.

To elaborate, you should carefully examine the details that suggest the front or rear of the object, such as eyes, nose, mouth, or tail, or tail lights. 
Additionally, you should closely check for features that imply a orientation, such as the orientation of the nose, whether one or both arms of a person are visible, or whether the wheels of a car appear as perfect circles, indicating left or right.

Instead of making overly general statements, you must create responses that are detailed enough to determine a single orientation. The answer should not just state that something is visible, but rather explain how these features suggest a particular orientation and provide specific details to justify the object's orientation.

 For example, you could write something like this:
 [Question]: Describe in detail the features in the image that indicate the orientation of the object.
 [Answer]: The two wheels of the car appear perfectly circular, and the car door is visible. This suggests that the object's orientation is either ``left" or ``right." However, since the headlights are on the right side of the image and the red taillights are on the left, the car's orientation is definitively to the ``right." 

 [Question]: Describe in detail the features in the image that indicate the orientation of the object.
 [Answer]: Both eyes of the person's face are visible, but one eye appears larger, and only one cheek is primarily visible, indicating a slanted orientation. The pointed part of the nose is closer to the left side of the image, and the left cheek is not clearly visible from the camera's perspective. Therefore, the person is facing ``front left"
 
    \vspace{0.2cm}
    Now your turn:
    +  context + ``[Question]: " + ``[Answer]:  " \\ \midrule

\textbf{Data Type3 Prompt} &

As a competent helper like Turn-by-turn navigation, you have the role of understanding the properties of the central object and creating common sense questions and corresponding answer appropriate for them.

You must use information in [Context]
If you make a question and answer, think carefully that matching the object's property and common sense with the probable action or capable happening.

But pay attention that while making question, you MUST not contain the orientation information directly you get from [Context] in question.

Alternatively you can represent with indirect word like corner of image, behind of it, away from camera's view point or something else.

Also, use natural expressions for orientation expressions like facing away from the camera, facing right else.

You must now generate a question, similar to the given examples, asking which orientation the object should be turned to face or turn away from the camera with the least angle of rotation, along with the corresponding answer.

I'll give you some example that you can reference.

For example, If you find the main object as a sport car, you can make a question like this:

[Question]: To make an object face the camera directly with the smallest rotation angle, in which direction should it turn? Choose from [clockwise, counterclockwise, flip, leave as is] and explain why.
[Answer]: The {sport car} is facing to the back, so you have to flip it.

[Question]: To make an object face the camera directly with the smallest rotation angle, in which direction should it turn? Choose from [clockwise, counterclockwise, flip, leave as is] and explain why.
[Answer]: The {sport car} is facing to the right, so you have to turn to the counterclockwise to look straight at the camera.

    \vspace{0.2cm}
    Now your turn:
    +  context + ``[Question]: ''  + ``[Answer]: ''  \\ \bottomrule

\end{tabular}%
}

\caption{Examples of training data for different data types.}
\label{tab:template_sample}
\end{table*}


\end{document}